\pdfoutput=1
\documentclass{article}

\usepackage{iclr2027_conference,times}

\usepackage{amsmath,amsfonts,bm}

\def\eqref#1{equation~\ref{#1}}

\def\1{\bm{1}}

\def\vzero{{\bm{0}}}

\def\mI{{\bm{I}}}

\DeclareMathAlphabet{\mathsfit}{\encodingdefault}{\sfdefault}{m}{sl}
\SetMathAlphabet{\mathsfit}{bold}{\encodingdefault}{\sfdefault}{bx}{n}

\newcommand{\E}{\mathbb{E}}

\newcommand{\R}{\mathbb{R}}

\usepackage{xcolor}
\usepackage{pifont}
\usepackage{amsmath}
\usepackage{amssymb}
\usepackage{booktabs}
\usepackage{array}
\usepackage[utf8]{inputenc}
\usepackage[english]{babel}
\usepackage{subcaption}
\usepackage{tikz}
\usetikzlibrary{calc}
\usepackage{graphicx}
\usepackage{multirow}
\usepackage{adjustbox}
\newcommand{\cmark}{\ding{51}}%
\newcommand{\xmark}{\ding{55}}%

\newcommand{\myparagraph}[1]{\vspace{0.1cm}\noindent {\bf #1}}
\newcommand{\eg}{\emph{e.g.}}
\newcommand{\appcref}[1]{Appendix~\cref{#1}}
\newcommand{\nr}[1]{\textit{#1}}

\usepackage{titletoc}
\newcommand{\myeq}{\mkern1.25mu{=}\mkern1.25mu}

\newcommand{\act}{\bm{A}}
\newcommand{\ctx}{\bm{E}}
\newcommand{\mem}{\bm{M}}
\newcommand{\stream}{\bm{X}}
\newcommand{\state}{\bm{s}}
\newcommand{\semb}{\bm{g}}
\newcommand{\noise}{\bm{\epsilon}}
\newcommand{\nreg}{n_r}
\DeclareMathOperator{\Enc}{Enc}
\DeclareMathOperator{\Pool}{Pool}
\DeclareMathOperator{\Attn}{Attn}

\usepackage[T1]{fontenc}
\usepackage{microtype}
\usepackage{url}
\usepackage{xurl}
\usepackage{hyperref}
\definecolor{linkblue}{HTML}{1A4D8C}
\hypersetup{colorlinks=true, linkcolor=linkblue, citecolor=linkblue, urlcolor=linkblue, pdfborder={0 0 0}, pdfborderstyle={}}
\usepackage[capitalise,nameinlink]{cleveref}
\crefformat{section}{\S#2#1#3}
\Crefformat{section}{\S#2#1#3}
\crefmultiformat{section}{\S#2#1#3}{ and~\S#2#1#3}{, \S#2#1#3}{ and~\S#2#1#3}
\crefname{equation}{Eq.}{Eqs.}
\crefname{appendix}{Appendix}{Appendices}

\title{REKI: Registering Knowledge in VLAs}

\author{Juliette Marrie$^{1}$, Paul Pacaud$^{2}$, Adrien Ramanana Rahary$^{3}$, Shizhe Chen$^{2}$, and Moritz B\"ohle$^{1}$\\
$^{1}$Kyutai $\cdot$ $^{2}$Inria $\cdot$ $^{3}$\'Ecole nationale des ponts et chauss\'ees}

\iclrfinalcopy
\begin{document}

\maketitle
\lhead{Preprint}

\begin{abstract}
We introduce REKI, a 2.4B vision-language-action model (VLA) pretrained exclusively on public data that sets a new Pareto frontier between performance and model size: on LIBERO-Plus, RoboTwin Clean2Rand and RoboCasa365, it is outperformed only by models about twice its size.
REKI consists of a vision--language model (VLM) followed by a flow-matching action expert, and is pretrained on a curated mixture of public robot, egocentric human and vision--language data.
Additionally, we endow it with a memory that gives the policy access to its recent past at negligible cost: a few register tokens in the action expert summarise the output of the VLM at each frame and persist in the expert's cache.
This memory improves success in simulation and on real robots, and makes oscillations between poses about four times rarer.
REKI uses a simple model design and training paradigm, and is optimised to be efficient at training and inference.
We release the full code across data curation, training, inference and evaluation, deployment, and all pretrained and finetuned model weights: \href{https://github.com/kyutai-labs/reki}{\texttt{github.com/kyutai-labs/reki}}.

\end{abstract}

\section{Introduction}
\label{sec:intro}

Vision-language-action models (VLAs) have become an effective paradigm for language-conditioned robot control.
A pretrained vision--language model (VLM) encodes camera views and instruction, based on which a smaller action expert predicts the next actions~\citep{black2024pi0, intelligence2025pi05, nvidia2025gr00tn1,shukor2025smolvla, yuan2026qwenrobot, ma2026internvla}.

The strongest VLAs today are pretrained partly on proprietary data, and neither these data nor the pretraining pipelines are released~\citep{intelligence2025pi05, yuan2026qwenrobot}, making it hard to reproduce their results. Since each new model changes architecture, data and training recipe at once, gains cannot be traced back to a single choice.
With REKI, we release the full pipeline, from data curation to deployment.
REKI has a simple design, is pretrained on public data only and sets a new Pareto frontier between performance and model size.
It pairs a VLM with a flow-matching action expert, and a memory mechanism gives the policy access to its recent past.

Preserving past context can improve control~\citep{octo, liu2025rdt1b, chi2025diffusion}, yet most VLAs predict actions from the current observation alone~\citep{kim2024openvla, black2024pi0, intelligence2025pi05, nvidia2025gr00tn1, shukor2025smolvla, molmoact, spatialvla, rdt2}.
Existing ways of incorporating such context, however, often introduce substantial overhead (\cref{fig:mechanisms}): feeding past frames to the VLM adds their image tokens to every VLM pass~\citep{jang2025contextvla, yuan2026qwenrobot}, recurrent methods require sequential training~\citep{cherepanov2026muvla, jiang2026robottt}, and others introduce additional, dedicated memory modules to the policy~\citep{shi2026memoryvla, koo2025hamlet}.

\begin{figure}[t]
    \centering
    \includegraphics[width=\linewidth]{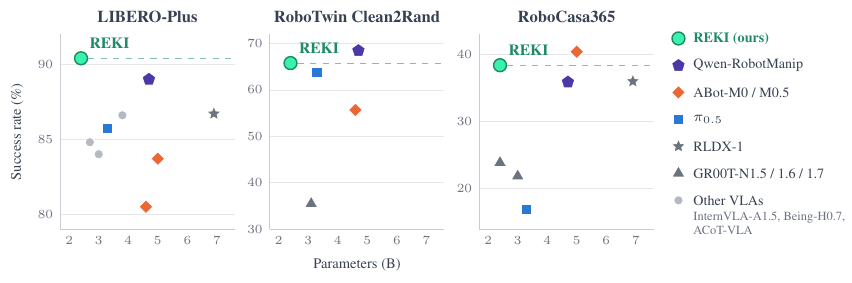}
    \caption{\textbf{Success rate against model size.} REKI sets a new Pareto frontier between performance and model size: it scores highest on LIBERO-Plus, and is only outperformed by models about twice its size on RoboTwin Clean2Rand and RoboCasa365. We report full results in \cref{fig:ood-size}.}
    \vspace{-.1cm}
    \label{fig:teaser}
\end{figure}

REKI summarises past context through a small set of registers~\citep{darcet2024registers} in the action expert.
The registers of each frame are cached across chunks (\cref{fig:expert}).
Each frame is thus encoded only once and no separate module, sequential unroll during training or test-time update mechanisms are necessary.
This yields clear gains at almost no overhead, and our 2.4B model REKI ranks first among VLAs pretrained on public data on LIBERO-Plus, RoboTwin Clean2Rand and RoboCasa365 (\cref{fig:teaser}), and generalises to real robots.

Our ablations show that memory consistently improves performance, with the most pronounced gains for the first past frame (\cref{fig:reki-memory}).
Beyond task success, we find that memory stabilises action generation: without memory, the policy often oscillates between poses across action chunks, while with memory, such oscillations are about four times rarer (\cref{fig:oscillation}).

\textbf{Our contributions are thus three-fold}: \textbf{(i)} We introduce REKI, a 2.4B-parameter VLA pretrained on public data, which sets a new Pareto frontier between performance and model size. \textbf{(ii)} We introduce register tokens in the action expert to act as a memory, which can be cached efficiently, and analyse how this reduces the failure modes of the memoryless policy. We hope REKI serves as a strong and accessible baseline for future VLA research. To foster VLA research, we release the model weights and our optimised training and inference code, from data curation to deployment on a real robot.

\section{Related work}
\label{sec:related}

\myparagraph{Open pretraining.} Many VLAs release their weights, and often their finetuning code, but not the code or the exact data mixture used for pretraining~\citep{openpi2025, nvidia2025gr00tn1, rdt2, zheng2025xvla, chen2026abotm05}.
The strongest VLAs are pretrained partly on proprietary data, collected~\citep{intelligence2025pi05, wang2026qwenvla, rdt2} or synthesised~\citep{yuan2026qwenrobot}, and several have no public weights~\citep{intelligence2026pi07, wang2026qwenvla, yuan2026qwenrobot}.
OpenVLA, Octo and RDT-1B released their full pipeline on public data such as Open X-Embodiment~\citep{kim2024openvla, octo, liu2025rdt1b}, followed by SmolVLA, MolmoAct and MolmoAct2, which also releases bimanual data collected by its authors~\citep{shukor2025smolvla, molmoact, fang2026molmoact2}.
These releases have helped the research community pretrain models again, change their data or design, and compare them under the same conditions.
Models pretrained on larger in-house data, with more recent design choices, have nonetheless kept a lead on most benchmarks.
Building on this line of work, REKI is pretrained on a broad mixture of public robot, egocentric human and vision--language data, and released in full, from data curation to deployment on a real robot.
It reaches the best results among VLAs pretrained on public data on three benchmarks (\cref{sec:results}), and is optimised for fast inference (\cref{sec:latency}).

\begin{figure}[t]
    \centering
    \includegraphics[width=\linewidth]{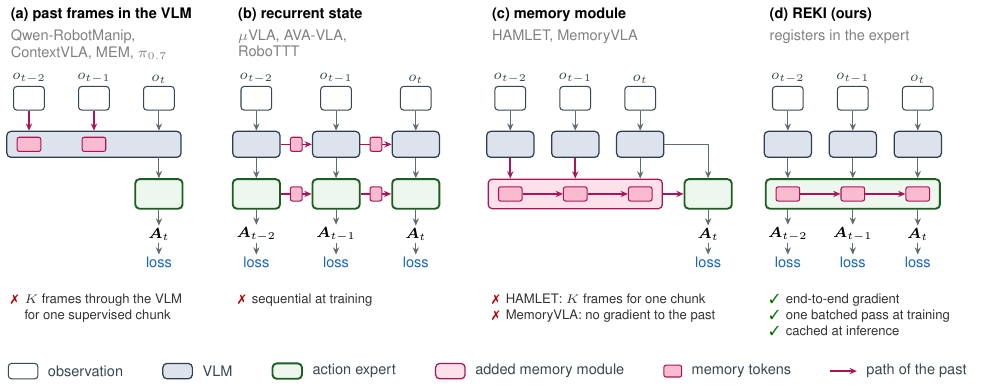}
    \caption{\textbf{How VLAs carry the past.} Each panel shows one training window of three frames, with a loss under each frame that receives one.
    \textbf{(a)} Past frames enter the VLM with the current one, in full~\citep{yuan2026qwenrobot}, compressed~\citep{torne2026mem, intelligence2026pi07} or from a cache~\citep{jang2025contextvla}.
    The VLM takes the whole window as input, and only the last frame is supervised.
    \textbf{(b)} A state is written at each step and read at the next, in the backbone~\citep{cherepanov2026muvla, xiao2026avavla} or as fast weights in the head~\citep{jiang2026robottt}.
    Each step needs the output of the previous one, so the window is unrolled step by step.
    \textbf{(c)} Each frame is encoded on its own, and an added module mixes the memory tokens of past frames into the prediction for the current one.
    HAMLET~\citep{koo2025hamlet} supervises the last frame only, and MemoryVLA~\citep{shi2026memoryvla} every frame, with the bank entries stored without gradient.
    \textbf{(d)} In REKI, the memory tokens are registers in the action expert.
    Their keys and values are kept across frames and read by the self-attention already present.
    The window is one batched pass, and every frame receives a loss. }
    \label{fig:mechanisms}
\end{figure}

\myparagraph{Memory in VLAs.} Most of these models act on the current observation alone, yet past context helps a policy identify the task state and recover from failed attempts~\citep{shi2026memoryvla, torne2026mem, gatedmemory}, and methods differ in where they pay for it (\cref{fig:mechanisms}).
The most direct approach stacks or compresses past frames in the VLM~\citep{yuan2026qwenrobot, torne2026mem, intelligence2026pi07, jang2025contextvla}.
Its cost then grows with the window, at least in the vision encoder~\citep{koo2025hamlet}.
A recurrent state~\citep{cherepanov2026muvla, xiao2026avavla} or fast weights~\citep{jiang2026robottt} keep the input short, but must be unrolled step by step in training.
Others store past features or keyframes in an extra module~\citep{shi2026memoryvla, koo2025hamlet, mark2026bpp, memer}.
Like $\mu$VLA and HAMLET, REKI carries the past in a few learnable tokens, but as registers~\citep{darcet2024registers} in the action expert.
The VLM encodes each frame once, a training window is one batched pass, and at inference memory adds under 1\% to the prediction time of a model without memory ($K\myeq1$, \cref{app:latency}).
Without memory, a policy can also switch between modes at each new action chunk~\citep{longcontextdp, lian2026intentvla}, which is attributed to perceptual aliasing in reinforcement learning~\citep{whitehead1991learning, singh1994learning, mccallum1996reinforcement}.
In REKI, memory makes such switches about four times rarer (\cref{sec:ablation}), without the extra policy queries of temporal ensembling or bidirectional decoding~\citep{act, liu2024bid}.

\newpage

\section{Background}
\label{sec:context}

\myparagraph{Problem formulation.} At each decision step $t$ we predict the chunk $\act_t \in \R^{H \times d_a}$ of the next $H$ actions of an embodied system, conditioned on the \emph{current context} $c_t$ with
\begin{equation}
    c_t = (z_t,\, o_t,\, \state_t),
    \label{eq:context}
\end{equation}
which gathers a text prompt $z_t$, camera observations $o_t$ and the current robot state $\state_t \in \R^{d_s}$.
More generally, a policy may condition on every context observed so far, $c_{1:t}$.

The camera observations $o_t = \{o_t^{1}, \dots, o_t^{V}\}$ gather the $V$ available views, static or embodied.
The robot state $\state_t$ is expressed as \emph{absolute joint angles} or an \emph{absolute end-effector pose}, and the actions as \emph{absolute joint angles} or \emph{delta end-effector poses} and yield the next $H$ states at a given control frequency.
Action and state dimensions ($d_a$, $d_s$) vary across embodiments and control modes.

\subsection{Diffusion policy}

Following prior work~\citep{chi2025diffusion, black2024pi0}, the action expert generates the chunk by flow matching~\citep{lipman2023flow}.
Write $\act = \act_t$ for the chunk generated at step $t$ and $c = c_{1:t}$ for the contexts observed up to it; the index $t$ is fixed in this subsection and omitted.
Given noise $\noise \sim \mathcal{N}(\vzero, \mI)$ and a flow time $\tau \sim p(\tau)$, the interpolant $\act^{\tau} = (1 - \tau)\,\act + \tau\,\noise$ moves from the clean chunk at $\tau = 0$ to noise at $\tau = 1$, and the expert $v_\theta$ is trained to regress its constant velocity $\noise - \act$,
\begin{equation}
    \ell_{\mathrm{fm}}(\theta;\, \act, c) =
    \E_{\tau, \noise}\!\left[
        \bigl\lVert v_\theta(\act^{\tau}, \tau \mid c) - (\noise - \act)
\bigr\rVert_2^2
    \right].
    \label{eq:fm-loss}
\end{equation}
The squared error is averaged over the valid steps and action dimensions of the chunk, so that every embodiment has the same weight whatever its number of action dimensions.
The action loss is the expectation of this term over the dataset $\mathcal{D}$ of (context, chunk) pairs,
\begin{equation}
    \mathcal{L}_{\mathrm{act}}(\theta) =
    \E_{(c, \act) \sim \mathcal{D}}\!\left[\ell_{\mathrm{fm}}(\theta;\,
\act, c)\right].
    \label{eq:act-loss}
\end{equation}
At inference, $\act^{1} \sim \mathcal{N}(\vzero, \mI)$ is integrated from $\tau = 1$ to 0 in ten Euler steps.

\subsection{Conditioning}
\label{sec:cond}

The action expert is conditioned on the camera observations, the text instruction and the robot state, through two separate paths.
The observations and the instruction are variable-length token sequences.
An encoder embeds them into an \emph{encoded context}
\begin{equation}
    \ctx_t = \Enc_\phi(z_t,\, o_t) \in \R^{L_t \times d},
    \label{eq:cond-context}
\end{equation}
whose tokens the expert reads as keys and values in its cross-attention blocks.
The state is a single low-dimensional vector.
It is pooled into one \emph{state embedding} $\semb_t = \Pool_\psi(\state_t) \in \R^{d_e}$, which modulates the expert through its adaptive normalisation~\citep{peebles2023dit}, alongside the flow-time embedding.
Since $\semb_t$ does not depend on $\tau$, it is computed once per frame for all Euler steps.

\myparagraph{Encoded context.} Following recent practice~\citep{yuan2026qwenrobot, yang2026abot}, $\ctx_t$ is the final-layer hidden states of a VLM, read by cross-attention.
This keeps the encoder and the action expert independent, and is reported to match or outperform alternatives such as processing the encoded context jointly with the noisy chunk by self-attention~\citep{wang2026qwenvla}, or attending the backbone at single or all layers~\citep{black2024pi0, intelligence2025pi05, shukor2025smolvla, nvidia2025gr00tn1}.
We use Qwen3.5-2B~\citep{qwen35} and pack the prompt and the camera views into a single stream (\appcref{fig:vlm}).
The prompt $z_t$ carries the language instruction, the embodiment, the control mode, the control frequency and a binned episode length (\cref{app:model}).
The embodiment, control-frequency and episode-length fields are dropped with probability 0.15, 0.2 and 0.2 during training.
Each view $o_t^v$ is preceded by a learned camera-role token (\texttt{side}, \texttt{overhead}, \texttt{wrist left}, \texttt{wrist right}) and embedded by the VLM's visual encoder at native resolution. Views above $448^2$ pixels are first downscaled with their aspect ratio preserved.
Views are shuffled and randomly dropped during training.

\myparagraph{State embedding.} $\Pool_\psi$ is one linear layer with a GELU per state layout, followed by a shared zero-initialised projection at the expert's width.
Its output modulates the expert through the adaptive normalisation.
The state is dropped with probability 0.2 during training.

\section{REKI model}
\label{sec:model}

Building on the policy of \cref{sec:context}, REKI gives the action expert a memory of prior frames through registers (\cref{sec:arch}), which are trained end to end with the flow-matching objective and cached at inference (\cref{sec:trainval}).

\subsection{Preserving past context}
\label{sec:arch}

Standard VLAs condition on the current context $c_t$ alone.
We give the expert access to the preceding frames through $\nreg$ memory tokens per frame, implemented as registers~\citep{darcet2024registers}.
The registers of frame $t$ cross-attend the encoded prompt and views of that frame, self-attend the registers of the earlier frames, and persist as keys and values in the expert's cache, implemented as a rolling window over the $K-1$ preceding frames with $K\myeq3$ (\cref{sec:trainval}).
Since they do not attend the action tokens and skip the adaptive normalisation, they depend only on the observations and can be shared across frames and denoising steps.

\myparagraph{Cost and consistency.} The memory is added without changing the encoder, which still processes one frame at a time, so the encoder cost is that of a memoryless policy.
During training, the frames are encoded in a single batched pass with no recurrence to unroll, and each frame is supervised on its own chunk~\citep{cherepanov2026muvla, jiang2026robottt}, so a window of $K$ frames yields $K$ training signals rather than one (\cref{fig:mechanisms}).

\myparagraph{Registers.} As in Qwen-RobotManip~\citep{yuan2026qwenrobot}, $\nreg$ learnable register tokens $\mem \in \R^{\nreg \times d_e}$ are prepended to the noisy actions in the expert's stream, so that frame $t$ enters it as $\stream_t = [\,\mem\,;\,\act_t^{\tau}\,]$, with the actions linearly projected to the expert's width $d_e$; we write $\mem_t$ for the register tokens of frame $t$.
While Qwen-RobotManip discards the registers after each forward pass, we keep their keys and values across frames, giving the action expert access to past observations.

\begin{figure}[t]
    \centering
    \begin{minipage}[c]{0.5875\linewidth}
        \centering
        \begin{subfigure}{\linewidth}
            \centering
            \begin{tikzpicture}[remember picture]
                \node[inner sep=0] (panelA)
                     {\includegraphics[width=\linewidth]{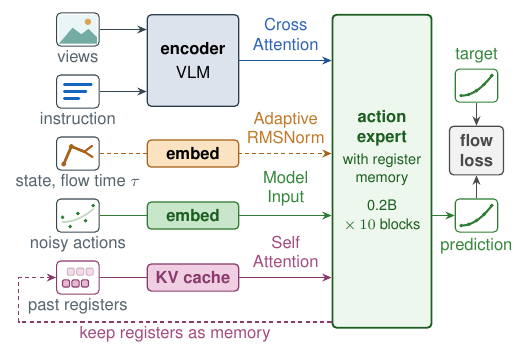}};
            \end{tikzpicture}
            \caption{REKI architecture.}
            \label{fig:model-a}
        \end{subfigure}
    \end{minipage}\hfill
    \begin{minipage}[c]{0.3825\linewidth}
        \centering
        \begin{subfigure}{\linewidth}
            \centering
            \begin{tikzpicture}[remember picture]
                \node[inner sep=0] (panelB)
                     {\includegraphics[width=\linewidth]{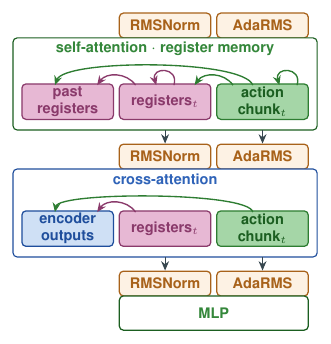}};
            \end{tikzpicture}
            \caption{One expert block.}
            \label{fig:model-b}
        \end{subfigure}
    \end{minipage}%
    \begin{tikzpicture}[remember picture, overlay]
        \definecolor{zoomc}{HTML}{4A4A4A}
        \tikzset{zoomline/.style={zoomc!75, line width=.8pt,
                                  dash pattern=on 2.6pt off 1.8pt,
                                  line cap=round, line join=round}}
        \coordinate (zxr) at ($(panelA.north west)!0.8081!(panelA.north east)$);
        \coordinate (zy1) at ($(panelA.north west)!0.0408!(panelA.south west)$);
        \coordinate (zjx) at ($(panelA.east)!0.82!(panelB.west)$);
        \draw[zoomline] (zxr |- zy1) -- (zjx |- zy1);
        \draw[zoomline] (zjx |- panelB.north) -- (zjx |- panelB.south);
        \draw[zoomline] (zjx |- panelB.north) -- (panelB.north west);
        \draw[zoomline] (zjx |- panelB.south) -- (panelB.south west);
        \coordinate (zmg) at ($(zxr |- zy1)!0.72!(zjx |- zy1)$);
        \coordinate (zmgc) at ($(zmg)+(0.12,0.32)$);
        \begin{scope}[shift={(zmgc)}]
            \fill[white] (0,0) circle (0.165);
            \draw[zoomc!80, line width=.8pt] (0,0) circle (0.165);
            \draw[zoomc!80, line width=1.1pt, line cap=round]
                 (-0.12,-0.12) -- (-0.25,-0.25);
        \end{scope}
    \end{tikzpicture}
    \caption{\textbf{REKI.} \textbf{(\subref{fig:model-a})} Camera views and instruction are encoded as context $\ctx_t$, which every expert block reads by cross-attention.
    The state embedding $\semb_t$ and the flow time modulate the actions through adaptive normalisation.
    The noisy chunk enters the expert's stream, and the flow loss compares the predicted velocity with its target.
    The register keys and values of past frames are kept in a cache and read by self-attention, which is the memory of \cref{sec:arch}.
    \textbf{(\subref{fig:model-b})} One of the ten blocks, holding $\nreg\myeq 32$ registers and the action tokens of the current frame.
    Arrows run from a query to what it reads.
    Action tokens read the current and past registers and themselves, and registers never read them. }
    \label{fig:expert}
\end{figure}

\myparagraph{Attention.} Every token of $\stream_t$ cross-attends $\ctx_t$.
In self-attention, writing $\Attn(\bm{Q};\,\bm{V})$ for attention with queries $\bm{Q}$ over the tokens of $\bm{V}$ as keys and values,
\begin{equation}
    \mem_t \leftarrow \Attn\bigl(\mem_t;\, [\mem_1; \dots; \mem_t]\bigr),
    \qquad
    \act_t^{\tau} \leftarrow \Attn\bigl(\act_t^{\tau};\, [\mem_1; \dots; \mem_t;\, \act_t^{\tau}]\bigr).
    \label{eq:regmem-mask}
\end{equation}
The history thus enters the velocity field through the attention support alone (\cref{fig:expert}b).

\myparagraph{Positions.} Self-attention positions are 2D, the frame index and the slot within the frame, encoded by the temporal and spatial axes of the VLM's multimodal rotary embedding (\cref{app:model}).

\subsection{Training and inference}
\label{sec:trainval}

\myparagraph{Objective.} A training sample is a window of $K$ frames from one episode, taken at a 1\,s stride and indexed from 1, the oldest, to $K$, the most recent.
At the start of an episode, a window holds fewer frames.
Each frame $t$ has its own chunk $\act_t$, and its contexts $c_{1:t}$ cover the frames up to it, without past states.
The loss averages the flow-matching loss of \cref{eq:fm-loss} over the chunks of all frames, each conditioned on the contexts up to its frame,
\begin{equation}
\mathcal{L}_{\mathrm{act}}(\theta) = \E_{(c_{1:K},\, \act_{1:K})\,\sim\,\mathcal{D}}\left[\frac{1}{K}\sum_{t=1}^{K} \ell_{\mathrm{fm}}\!\left(\theta;\, \act_t,\, c_{1:t}\right)\right].
\label{eq:regmem-loss}
\end{equation}
Since every later frame of the window reads the registers of a frame, these registers receive gradient from all of their losses.
At pretraining, the LM head of the VLM is also trained to predict the subtask when one is annotated and the answers of the vision--language data, with a cross-entropy loss of weight 0.1 (\appcref{fig:vlm}).

\myparagraph{Multi-draw variance reduction.}
\label{sec:multidraw}
The loss of \cref{eq:fm-loss} is an expectation over the flow time and the noise, and a single $(\tau, \noise)$ draw per frame gives a noisy estimate of it.
An extra draw costs one pass of the expert, whereas the context and the registers it reads cost one pass of the encoder, which dominates the cost of a training step.
As in \citet{yuan2026qwenrobot} and \citet{allshire2026abc}, we therefore estimate the expectation from eight independent draws per frame.
The eight noisy chunks share the encoded context and the registers of their frame, so the encoder runs once for all of them.
The registers of frame $s$ thus receive gradient from the eight draws of every frame from $s$ to $K$.

\myparagraph{Implementation.} The frames of a window are packed into a single batch and kept independent in the backbone by \texttt{cu\_seqlens} boundaries.
In the expert, the self-attention of \cref{eq:regmem-mask} is computed by one variable-length \textsc{FlashAttention} call.
For this, the keys and values that each group of queries reads are gathered into one sequence: the register tokens $\mem_{1:t}$ for the registers of frame $t$, and $\mem_{1:t}$ followed by the action tokens of one noisy chunk for the queries of that chunk (\appcref{fig:attn}).

\myparagraph{Inference.} At inference, each frame needs one encoder pass over the current observation and ten Euler steps on the action tokens.
The registers do not read the action tokens and skip adaptive normalisation (\cref{sec:arch}), so neither they nor the context depend on the flow time $\tau$ or on the state.
The registers of the current frame are thus computed together with the action tokens in the first Euler step, and their keys and values are reused at the nine others.
The register keys and values are also kept for the frames that follow, $(K-1)\,\nreg$ pairs per expert block.
The cache is a rolling window, so the registers of the oldest frame are dropped when those of a new frame enter (\appcref{fig:register-cache}).
At training, each window of $K$ frames is processed in isolation.
At inference, the cached registers were themselves computed with the registers of earlier frames, so information can persist beyond the window.
The VLM prefill and the expert are each compiled into a single CUDA graph.
With two $256{\times}256$ images and a chunk of ten actions, REKI predicts a chunk in 21.8\,ms on one H100 at $K\myeq3$, the lowest latency among the VLAs compared in \cref{sec:latency}.
In the measurement of \cref{fig:latency-main}, the VLM prefill takes 9.5\,ms and the Euler steps 8.1\,ms, and reading the cached registers adds 0.1\,ms.

\section{Experimental setup}

\subsection{Pretraining data}
\label{sec:data}

\myparagraph{Robot manipulation.} We train on a broad mixture of public teleoperation datasets: RoboSet~\citep{bharadhwaj2024roboset}, Galaxea~\citep{jiang2025galaxea}, AgiBot World 2026~\citep{agibotworld2026}, RoboCOIN~\citep{wu2025robocoin}, RoboMIND~\citep{wu2025robomind}, RDT-1B~\citep{liu2025rdt1b}, DROID~\citep{khazatsky2024droid}, BridgeData V2~\citep{walke2023bridgedata}, and twelve datasets of Open X-Embodiment~\citep{oxe2024} including RT-1~\citep{brohan2023rt1} and BC-Z~\citep{jang2022bcz} (\cref{app:embodiments}), complemented by motion-planned simulation from InternData-A1~\citep{tian2026interndata} and GR00T-X-Embodiment-Sim~\citep{nvidia2025gr00tn1} and bimanual YAM demonstrations from ABC~\citep{allshire2026abc}.
These span single-arm, bimanual, mobile, and humanoid platforms across tabletop, in-the-wild, and loco-manipulation settings (\appcref{tab:embodiments}).
Action and state dimensions, control conventions and units vary across sources, and a robot may appear under several conventions.
Following Qwen-VLA~\citep{wang2026qwenvla}, we keep each dataset's native control convention rather than remapping to a shared action space, normalise each action dimension to $[-1,1]$ by per-dataset quantiles, and zero-pad actions to a common width under a loss mask.
States are absolute, actions are absolute joint positions or delta end-effector poses, and rotations are harmonised to one Euler parameterisation.
The prompt and the per-layout state projections of \cref{sec:cond} tell the policy which convention a sample follows.
In the spirit of Qwen-RobotManip's multi-stage filtering~\citep{yuan2026qwenrobot}, we curate every source before training: we keep only the degrees of freedom the policy commands, discard static and near-zero-variance episodes, and correct per-dataset sign conventions.
\Cref{app:curation} gives the curation stages and the action representation.

\myparagraph{Egocentric human manipulation.} We add three egocentric human manipulation datasets, EgoDex~\citep{hoque2026egodex}, EgoVerse~\citep{punamiya2026egoverse} and Xperience~\citep{ropedia2026xperience10m}, recorded from head-mounted or hand-held cameras.
They contribute ${\sim}350$k episodes and 170M frames.
The action target follows Qwen-VLA's 32D bimanual layout~\citep{wang2026qwenvla}, wrist motion and hand articulation per hand, here with the wrist expressed in the camera frame; no proprioceptive state is used, and each is resampled to the pace of robot teleoperation~\citep{yuan2026qwenrobot}.
\Cref{app:ego} gives the sources, target definition and resampling.

\myparagraph{Vision--language data.} We train on the FineVision~\citep{wiedmann2025finevision} and LLaVA-OneVision-1.5~\citep{an2025llavaov15} datasets most relevant for spatial reasoning (\cref{app:vldata}), as well as targeted spatial-grounding and pointing corpora: RoboPoint~\citep{yuan2024robopoint}, RefSpatial~\citep{zhou2025refspatial}, and PixMo~\citep{deitke2024pixmo}.
This auxiliary supervision preserves the backbone's perception and spatial reasoning, and is mixed in at a small ratio with the robot and egocentric data.

\subsection{Models}
\label{sec:setup}

REKI pairs the Qwen3.5-2B vision--language backbone with an action expert of 10 blocks of width 768, 32 registers per frame and a memory window of $K\myeq3$ frames.
It is pretrained on the mixture above for 200k steps on 32 H100 GPUs with a global batch of 2{,}880 samples, then finetuned on each benchmark separately.
The backbone is trained at $0.1\times$ the learning rate of the expert, and 10\% of the training samples are vision--language data.
Finetuning steps, node counts and the number of evaluation rollouts per task are in \appcref{tab:setups}, the remaining hyperparameters in \appcref{tab:hparams}.
Rows marked \emph{scratch} skip the pretraining and finetune the initialised model directly on the benchmark.\footnote{Baseline numbers are taken from published results (\appcref{tab:provenance}).
The training budgets published for these baselines are in \appcref{tab:training-budgets-samples}.
Models in italics in the tables are not released at the time of writing.}

\subsection{Visual data augmentation}
\label{sec:augmentation}

We apply data augmentation to make the model robust to visual variations: colour jittering, mild random cropping, random camera dropout and shuffled camera order.
When finetuning on RoboTwin, RoboCasa365 and RoboCasa-GR1, we add a mirroring augmentation, which extends the horizontal flip common in image classification to robot data.
With probability 0.5, a sample is replaced by its left-right mirror image.
The images are flipped, the state and the actions are reflected through the same plane, the arms of a two-arm robot are swapped, and the prompt marks the sample as mirrored (\cref{app:mirroring}).
The augmentations and other training settings, such as the state dropout rate, have a large effect on robustness: on LIBERO-Plus, disabling camera dropout or the image augmentations lowers the success under camera-viewpoint changes by more than 20 points, and disabling state dropout lowers the success under changes of the initial robot state by 8.1 points (\cref{app:ablations}).

\section{Experiments}
\label{sec:evaluation}

\subsection{Results}
\label{sec:results}

\begin{figure}[t]
  \centering
  \includegraphics[width=\linewidth]{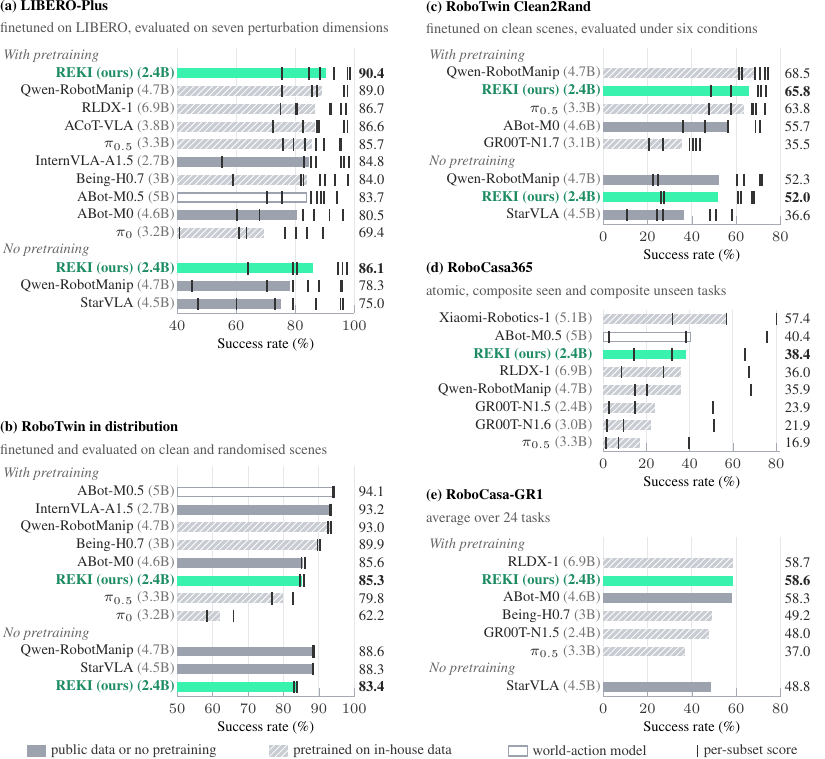}
  \caption{\textbf{Benchmark results.} Out of distribution, on LIBERO-Plus (a) and RoboTwin Clean2Rand (c), REKI leads the compared VLAs pretrained on public data; in distribution, on RoboTwin (b), RoboCasa365 (d) and RoboCasa-GR1 (e), it is competitive with larger models.
  Hatched bars mark models pretrained on in-house data and outlined bars world-action models.
  Each bar gives the overall score and the ticks the subset scores. 
  Full results for (a) to (d) are in \appcref{tab:libero-plus,tab:robotwin-c2r,tab:robocasa365}.}
  \label{fig:ood-size}
  \begingroup
  \phantomsubcaption\label{fig:ood-lp}
  \phantomsubcaption\label{fig:robotwin-id}
  \phantomsubcaption\label{fig:ood-c2r}
  \phantomsubcaption\label{fig:robocasa}
  \phantomsubcaption\label{fig:gr1}
  \endgroup
\end{figure}

\myparagraph{Benchmarks.} LIBERO~\citep{liu2023libero} has four suites of ten single-arm tabletop tasks, and LIBERO-Plus~\citep{fei2025liberoplus} perturbs its 40 tasks along seven dimensions, from the camera viewpoint and the initial robot state to lighting and object layout.
The models are finetuned on the four suites and evaluated on both.
RoboTwin 2.0~\citep{chen2025robotwin2} has 50 tasks on the dual-arm AgileX ALOHA, in clean scenes and in scenes with a randomised background, lighting, clutter and table height.
For the out-of-distribution evaluation (Clean2Rand;~\citealp{yuan2026qwenrobot}), the models are finetuned on clean scenes only and evaluated under six conditions: on clean scenes (Easy), under each randomisation alone and under all four together (Hard).
The Clean2Rand score is the mean over these six conditions.
RoboCasa365~\citep{nasiriany2026robocasa365} has kitchen tasks for a Franka arm on a mobile base.
The models are trained on the human demonstrations of its 300 pretraining tasks and evaluated on 50 target tasks: 18 atomic tasks, 16 composite tasks seen in training and 16 composite tasks unseen in training.
RoboCasa-GR1~\citep{nvidia2025gr00tn1} has 24 tabletop tasks for the Fourier GR-1 humanoid, with 1000 demonstrations per task.
We compare mainly with VLAs of similar complexity, which pair a VLM with an action expert, such as $\pi_{0.5}$, GR00T, ABot-M0 and InternVLA-A1.5.
As a gauge of the current upper bound on these benchmarks, we also report larger models pretrained on in-house data, Qwen-RobotManip, RLDX-1 and Xiaomi-Robotics-1, and the world-action model ABot-M0.5, built on a video generation model.
They are trained under different conditions, with more or proprietary data or a video model, so they are a reference rather than a like-for-like comparison.
\Cref{fig:ood-size} summarises the results across benchmarks, and \appcref{tab:libero-plus,tab:robotwin-c2r,tab:robocasa365} list them in full.

\myparagraph{Out of distribution.} REKI stands out most when the test conditions differ from training.
On LIBERO-Plus and Clean2Rand, it ranks first among the VLAs pretrained on public data, despite having the fewest parameters (2.4B, \cref{fig:ood-lp,fig:ood-c2r}).
It leads the next of these models by 5.6 points on LIBERO-Plus (InternVLA-A1.5, 84.8) and by 10.1 points on Clean2Rand (ABot-M0, 55.7).
On the composite tasks of RoboCasa365 unseen in training, it is on par with Qwen-RobotManip (14.1 against 14.9) and behind Xiaomi-Robotics-1 (32.1), both larger and pretrained on in-house data.

\myparagraph{In distribution.} When trained and evaluated on the same tasks, REKI is competitive with larger models.
On RoboCasa365, it reaches 38.4, ahead of the larger RLDX-1 (36.0) and Qwen-RobotManip (35.9), both pretrained on in-house data (\cref{fig:robocasa}), and on RoboCasa-GR1 it reaches 58.6, on par with the larger RLDX-1 (58.7) and ABot-M0 (58.3) (\cref{fig:gr1}).
On LIBERO, it reaches 97.5, where most models lie between 97.6 and 99.4.
REKI is weakest on RoboTwin, where it reaches 85.9 and 84.6, below the similarly sized InternVLA-A1.5 (93.3 / 93.0), which is supervised by an additional video generation model during training and uses a more complex, multi-stage pretraining (\cref{fig:robotwin-id}).

\myparagraph{Pretraining gains.} Trained from scratch on each benchmark, REKI reaches 52.0 on Clean2Rand, on par with Qwen-RobotManip (52.3) and ahead of StarVLA (36.6), which both have larger VLMs (4.5B and 4.4B).
REKI is ahead of both under the Hard condition (26.3 against 22.6 and 10.6).
On LIBERO-Plus it leads with 86.1, ahead of Qwen-RobotManip (78.3) and StarVLA (75.0).
Pretraining on public data benefits REKI most out of distribution, and in particular under the RoboTwin randomisations (\appcref{tab:robotwin-c2r,tab:libero-plus}).
In distribution, it adds 2.9 and 0.8 points on RoboTwin and 1.0 on LIBERO.
Out of distribution, it adds 13.8 points on Clean2Rand and 4.3 on LIBERO-Plus, and the gain grows where REKI without pretraining fails most, with 30.3 points under clutter and 22.4 under the Hard condition.
On LIBERO-Plus, the gain is also largest under perturbed robot states (11.7).

\myparagraph{Additional data or supervision.} REKI is outperformed by several larger and more recent models that add data or supervision to the vanilla recipe, with large in-house data collections~\citep{xiaomi2026robotics1} or video generation models~\citep{song2026uniwam, zhang2026prowam}.
Similarly, InternVLA-A1.5 uses the same VLM as REKI, with an additional video generation model, a larger action expert (24 layers vs 10 for REKI) and a multi-stage pretraining, and is ahead in distribution on RoboTwin (93.3 vs 85.9).
Out of distribution, it is behind on LIBERO-Plus (84.8 vs 90.4).
Qwen-RobotManip, pretrained with additional synthetic data, gains more from pretraining than REKI (16.2 vs 13.8 points on Clean2Rand).
REKI reaches its lead out of distribution without such additions, and we expect it to benefit similarly from additional data, be it real~\citep{xiaomi2026robotics1}, synthetic~\citep{yuan2026qwenrobot} or generated by a world model~\citep{kim2026rldx1, gigabrain2026gigabrain07}.

\subsection{Ablation}
\label{sec:ablation}

\begin{figure}[t]
  \centering
  \includegraphics[width=\linewidth]{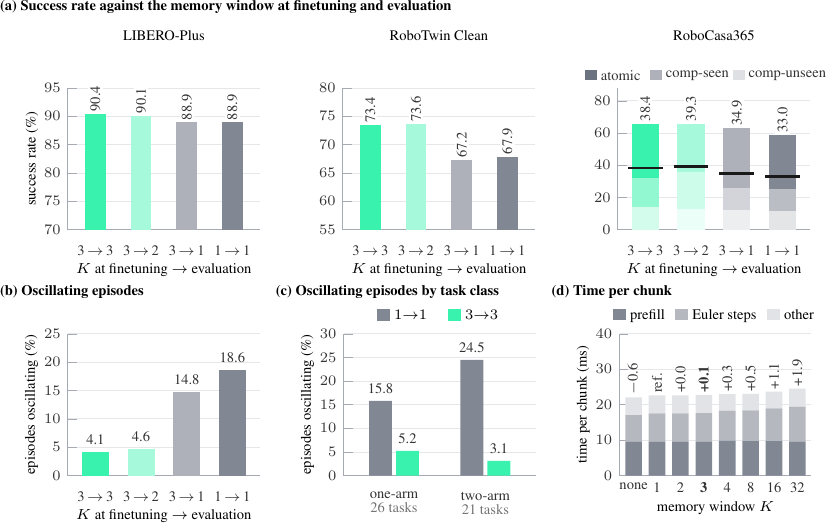}
  \caption{\textbf{Memory window, oscillation and inference cost.} REKI trained with $K\myeq3$ and evaluated with $K\myeq3,2,1$ ($3\to3$, $3\to2$, $3\to1$), and a memoryless model ($1\to1$). (a) On RoboCasa365, the black line gives the average. (b) RoboTwin Clean episodes in which an arm reverses its motion at three or more consecutive chunks (50 tasks, 20 matched seeds each). (c) Two-arm tasks move both arms in at least 80\% of the successful episodes with memory, one-arm tasks a single arm.
  The three other tasks are left out. (d) On one H100.
  The numbers give the difference to $K\myeq1$ in ms.}
  \label{fig:memory}
  \begingroup
  \phantomsubcaption\label{fig:reki-memory}
  \phantomsubcaption\label{fig:oscillation}
  \phantomsubcaption\label{fig:osc-class}
  \phantomsubcaption\label{fig:latency-main}
  \endgroup
\end{figure}

\myparagraph{Memory window.} \Cref{fig:reki-memory} varies the memory window of REKI at finetuning and at evaluation, on LIBERO-Plus after finetuning on LIBERO, on RoboTwin Clean and on RoboCasa365.
Finetuned and evaluated with three frames ($3\to3$), REKI reaches 90.4 on LIBERO-Plus, 73.4 on RoboTwin and 38.4 on RoboCasa365, against 88.9, 67.9 and 33.0 without memory ($1\to1$).
Most of the gain already comes with the first past frame ($3\to2$: 90.1, 73.6 and 39.3).
As every frame of a training window is supervised (\cref{eq:regmem-loss}), the model finetuned with memory also learns to act without it.
Evaluated without memory ($3\to1$), it performs close to the model finetuned at $K\myeq1$ (88.9, 67.2 and 34.9), so the gain comes mainly from reading the past frames at inference, not from training on windows.

\begin{figure}[t]
  \centering
  \includegraphics[width=\linewidth]{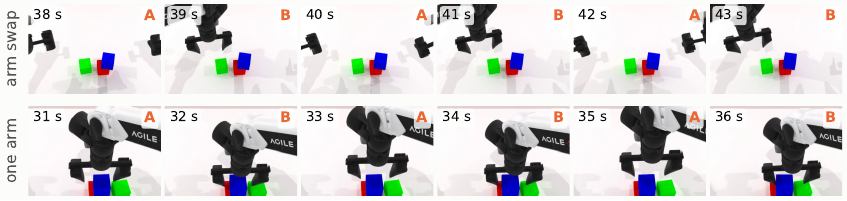}
  \caption{\textbf{Oscillation without memory.} Two RoboTwin rollouts of REKI trained without memory, stacking three blocks.
  Frames are one chunk apart (1\,s), and A and B mark the two poses the arm alternates between.
  In the top row, the two arms swap at every chunk, one reaching while the other retracts.
  In the bottom row, a single arm moves back and forth.
  The blocks do not move in either row.}
  \label{fig:filmstrip}
\end{figure}

\myparagraph{Oscillation.} For each arm, we compare how it moves during one chunk with how it moved during the previous one, ignoring small movements.
\Cref{fig:oscillation} counts the episodes in which an arm moves in roughly the opposite direction at three or more consecutive chunks, that is, it goes back and forth at least twice (\cref{fig:filmstrip}).
We set this criterion on 16 of the 50 tasks and confirmed the effect of memory on the other 34.
Fewer than 5\% of these episodes succeed.
With the same weights, removing the past frames at evaluation raises their share from 4.1\% to 14.8\%, and one past frame already brings it down to 4.6\%.
Trained without memory, REKI oscillates in 18.6\% of the episodes, and in half of those that fail.
The difference is largest in two-arm tasks (\cref{fig:osc-class}).

\myparagraph{Why memory helps.} A policy without memory maps the current observation to the next chunk.
If two poses A and B each look like demonstration states that continue towards the other, the chunk predicted at A leads to B, and the chunk predicted at B leads back to A. We attribute the oscillation to this form of perceptual aliasing, under which memoryless policies are known to fail~\citep{whitehead1991learning, singh1994learning, mccallum1996reinforcement}.
A short history resolves the ambiguity, since a window that has just come from B does not match demonstrations that move towards B. In two-arm tasks, the most frequent cycle is an arm swap, in which one arm reaches while the other retracts (65 against 1 of 1000 episodes without and with memory, top row of \cref{fig:filmstrip}). It typically starts a few seconds after a release.
Memory does not rescue all these episodes: where the policy without memory oscillates, the policy with memory succeeds on 45\% of the seeds.

\subsection{Inference latency}
\label{sec:latency}

\begin{figure}[t]
  \centering
  \includegraphics{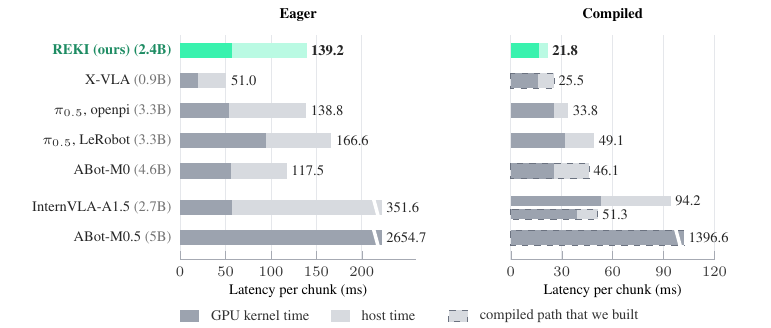}
  \caption{\textbf{Inference latency (H100)}, from the raw observation to the actions of one chunk, with LIBERO inputs, without compilation (left) and compiled (right).
  Dashed bars are compiled code that we built.
  ABot-M0.5 runs with its RoboCasa365 weights, the only ones released, and with its own LIBERO setting, which takes two $128{\times}128$ images.}
  \label{fig:latency-bars}
\end{figure}

\myparagraph{Protocol.} We measure the latency of one call, from the raw observation to the unnormalised actions of one chunk, including preprocessing and the transfers between host and device.
This latency bounds how quickly a policy can react to a new observation and how often it can replan, whatever the number of actions it predicts.
In practice, policies replan well before the end of a long chunk, typically in under a second, since executing more actions open loop makes them slower to react~\citep{black2024pi0, chi2025diffusion, liu2024bid}.
Each model predicts one chunk from two $256{\times}256$ images ($128{\times}128$ for ABot-M0.5), the robot state and a fixed prompt, at batch size one, with its released checkpoint and inference code.
Each value in \cref{fig:latency-bars} is the median over 320 chunks (80 for ABot-M0.5) and three repetitions, each on its own GPU of a node reserved for the timing (\cref{app:latency-runs}).

\myparagraph{Compilation.} Most of the compared models are released without a compiled path, or with one that can be further optimised.
Without compilation, all models are slow (\cref{fig:latency-bars}, left): REKI, $\pi_{0.5}$ with openpi~\citep{openpi2025} and ABot-M0~\citep{yang2026abot} take 118 to 139\,ms per chunk, InternVLA-A1.5~\citep{ma2026internvla} 352\,ms, and X-VLA (0.9B;~\citealp{zheng2025xvla}) 51\,ms.
In most of these runs, the GPU is busy less than half of the time, so compiling the prediction, which removes most of the host time, makes it two to seven times faster (\cref{fig:latency-bars}, right).
We compile REKI entirely into CUDA graphs, including its memory (\cref{app:latency}).
For a fair comparison, we also compile X-VLA, ABot-M0, InternVLA-A1.5 and ABot-M0.5~\citep{chen2026abotm05} without changing their computation, and report both their released path and ours.

\myparagraph{Results.} With compilation, REKI is the fastest of the compared models (\cref{fig:latency-bars}, right): it predicts a chunk in 21.8\,ms, against 25.5\,ms for X-VLA, 33.8\,ms for $\pi_{0.5}$ with openpi, 46.1\,ms for ABot-M0 and 51.3\,ms for InternVLA-A1.5.
Its GPU time per chunk, 16.5\,ms, is on par with that of X-VLA (16.1\,ms) and below that of the other models, which need at least 25.3\,ms.
InternVLA-A1.5 uses the same VLM as REKI, and the difference lies in its action expert, which runs at each of the ten Euler steps: with 0.5B parameters in 24 layers against 0.2B in 10, its Euler steps take 34.9\,ms against 8.2\,ms for REKI (\appcref{tab:latency-phases}).
ABot-M0.5 generates future video latents before it predicts actions, and takes 2.7\,s per chunk without compilation and 1.4\,s with the CUDA graphs that we built.

\section{Real robot experiments}

\begin{figure}[t]
\centering
\begin{minipage}[b]{0.3\linewidth}
\centering
\includegraphics[width=\linewidth]{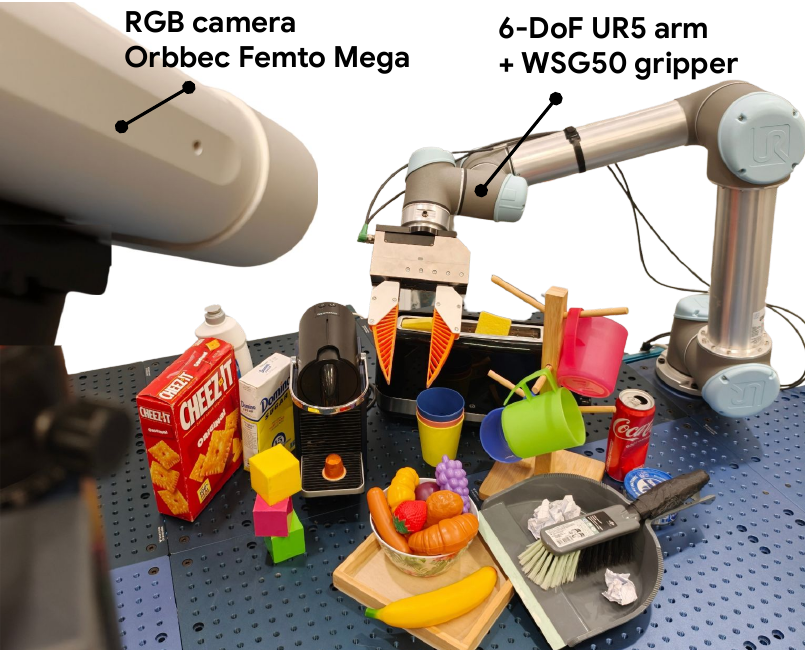}
\end{minipage}\hfill
\begin{minipage}[b]{0.68\linewidth}
\centering
\includegraphics[width=\linewidth]{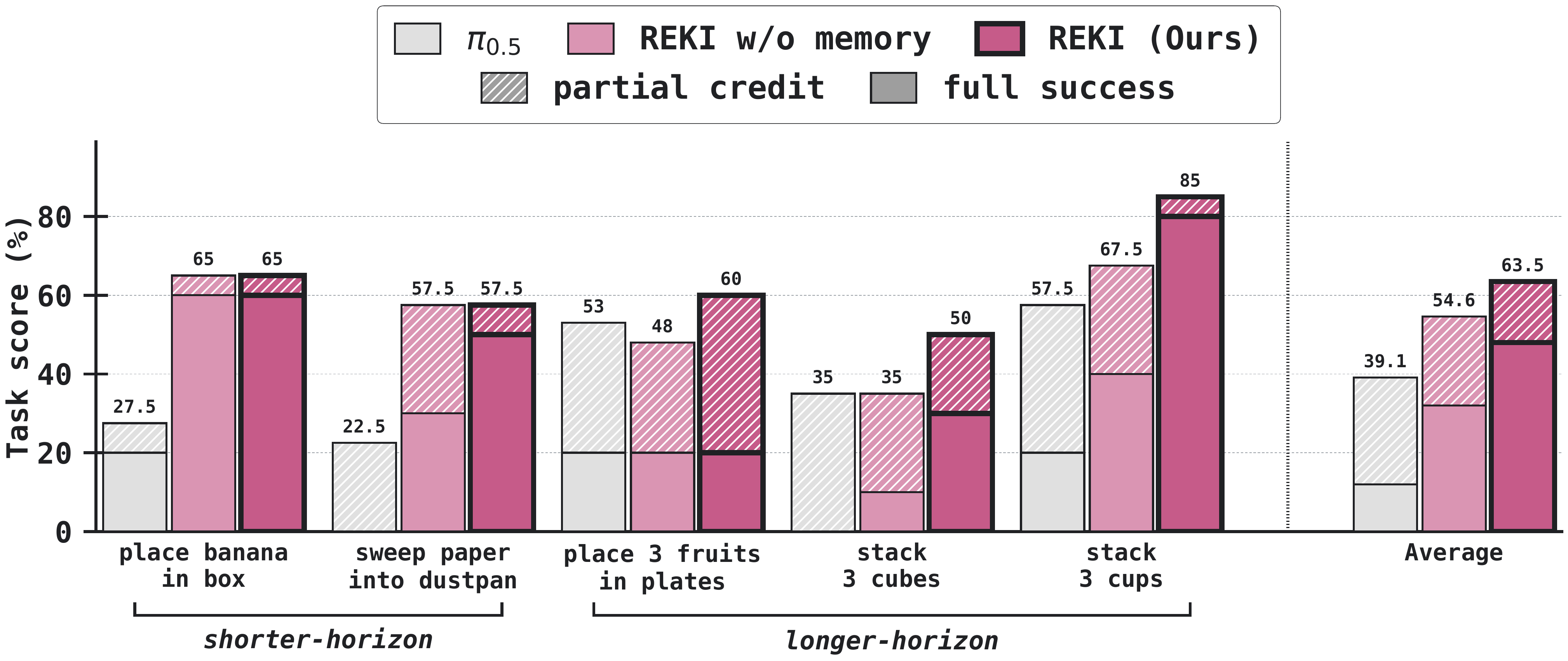}
\end{minipage}
\caption{\textbf{Real-robot experiments.} Left: the UR5 arm with a WSG50 gripper, observed by a fixed Orbbec Femto Mega RGB camera, with the task objects and distractors.
Right: task score of REKI with and without memory and of $\pi_{0.5}$~\citep{intelligence2025pi05} on two shorter- and three longer-horizon tasks, over 10 episodes each.
Solid bars give full successes, hatched bars partial credit.}
\label{fig:real-robot}
\vspace{-.2cm}
\end{figure}

\myparagraph{Experimental setup.} We use a 6-DoF UR5 robotic arm with a Weiss WSG50 gripper and UMI-style parallel jaws.
The action loop runs at 15 Hz.
The robot operates over a 1\,m${\times}$1\,m tabletop workspace observed by a fixed Orbbec Femto Mega RGB camera, see \cref{fig:real-robot}, left.
Observations include an RGB image, and a 7D absolute joint state.
Actions are specified as 7D absolute joint state.

\myparagraph{Tasks and protocol.} We evaluate performance across five manipulation tasks, and report task success rates and partial subtask success rates.
Tasks span diverse action primitives, including pick, place, stack, and sweep.
The tasks and partial success scores are defined in~\cref{app:real-robot}.
We collect 50 training episodes per task with a Meta Quest controller, and evaluate on 10 test episodes per task, with randomised initial object placements and distractors in each episode.

\myparagraph{Baselines.} We finetune the pretrained checkpoints of REKI and $\pi_{0.5}$ on the five tasks for 3{,}000 optimiser steps, with a global batch of 360 samples.
Both models use a learning rate of $5\times10^{-5}$, held constant after warmup.
The action chunk has 50 steps (3.3\,s) for $\pi_{0.5}$, fixed by its pretraining, and 30 steps (2\,s) for REKI.
In both models, only the last four vision encoder blocks are trained.
The backbone is trained at $0.1\times$ the learning rate of the action expert.
REKI is finetuned with a memory window of $K\myeq3$ and without memory ($K\myeq1$). $\pi_{0.5}$ has no memory.

\myparagraph{Results.} Across the five tasks, REKI outperforms $\pi_{0.5}$ by 7 to 37.5 percentage points (\cref{fig:real-robot}, right).
Removing the memory leaves performance on the shorter-horizon tasks mainly unchanged (65 against 65 and 57.5 against 57.5), but consistently degrades it on the longer-horizon tasks, where REKI outperforms its memoryless variant by 12 to 17.5 percentage points, supporting that memory matters most when tasks require tracking progress over many steps.
For qualitative examples, see \cref{fig:real-robot-qualitative}.

\section{Discussion}
\label{sec:discussion}

\myparagraph{A vanilla VLA.} REKI follows a standard VLA recipe: a VLM's last hidden states are read by a flow-matching action expert, pretrained in a single stage on public robot, egocentric human and vision--language data, and then finetuned on each benchmark.
It uses no proprietary data~\citep[cf.][]{yuan2026qwenrobot, kim2026rldx1}, no world model~\citep[cf.][]{chen2026abotm05} and no additional training stages.
Even so, it can compete with, and often outperform, VLAs of similar size pretrained on in-house data, and it leads the compared VLAs pretrained on public data out of distribution (\cref{fig:ood-size}), where pretraining adds 13.8 points on Clean2Rand and 4.3 on LIBERO-Plus (\cref{sec:results}).
Many of the strongest entries on current leaderboards take the other route: they are world-action models built on video generation models~\citep{chen2026abotm05, wang2026patchwam, zhang2026prowam, song2026uniwam} or VLAs pretrained on large proprietary data~\citep{xiaomi2026robotics1, gigabrain2026gigabrain07}.
These models set the current state of the art, but they can be costly to run (\cref{sec:latency}).
Their pretraining is also hard to reproduce, which makes it difficult to separate the effects of data scale, model size, modelling choices and additional supervision such as world modelling.
We intend REKI to serve as an open reference on which each of these factors can be tested in isolation.

\myparagraph{An efficient open baseline.} We release REKI with its full pipeline, from data curation to deployment, and the checkpoints of every benchmark, so that new architectures, data mixtures or training objectives can be pretrained and compared against it under fixed conditions.
The simplicity of the recipe and the efficiency of the code keep such comparisons affordable: finetuning REKI takes 8 H100 GPU-hours on our real-robot tasks and 40 to 400 on the benchmarks (\cref{tab:real-robot-hparams,tab:setups}).
The memory adds no cost at training (\cref{sec:trainval}) and less than 1\% of the inference time, so REKI is an efficient starting point for studying memory in embodied AI.
In the released code, REKI runs entirely as CUDA graphs, memory included, and predicts an action chunk from the LIBERO inputs in 21.8\,ms on one H100.
It stays the fastest of the compared models even when their released code is compiled (\cref{sec:latency}).

\myparagraph{Why inference speed matters.} Inference speed matters not only at deployment, where it bounds how quickly a policy can react (\cref{sec:latency}), but also for training.
VLAs are increasingly post-trained with reinforcement learning, in simulation~\citep{li2025simplevlarl, chen2025pirl, zang2026rlinfvla} or in learned world models~\citep{zhu2025wmpo, sharma2026worldgymnast}, where the policy is queried for every action chunk of every rollout, so that its inference time is multiplied across the whole training run.
At 21.8\,ms per chunk, REKI offers a strong open starting point for such post-training and lowers the barrier to it, a step towards making it accessible beyond the few groups with large compute budgets.

\myparagraph{Memory.} In REKI, a few register tokens in the action expert summarise each frame and stay in the expert's cache, so the VLM still sees each frame only once.
Even this short memory helps: with three frames, success rises by 1.5 to 5.5 points on LIBERO-Plus, RoboTwin and RoboCasa365, and by 12 to 17.5 points on the longer real-robot tasks.
Most of the gain comes from a single past frame, which in particular reduces repetitive oscillations between two poses, suggesting that a little temporal context is enough to tell apart observations that look alike (\cref{sec:ablation}).
A memory of a few seconds therefore appears sufficient for most tasks in standard benchmarks.
Whether the registers can also carry the much longer context of tasks that depend on events far in the past remains to be tested.

\newpage
\bibliography{main}
\bibliographystyle{iclr2027_conference}

\clearpage
\appendix
\section*{Appendix}
\suppressfloats[t]
\startcontents[appendix]
\printcontents[appendix]{}{1}{\setcounter{tocdepth}{2}\setlength{\parskip}{0pt}%
  \titlecontents{section}[1.5em]{\addvspace{2pt}\bfseries}%
    {\contentslabel{1.5em}}{\hspace*{-1.5em}}{\hfill\contentspage}%
  \titlecontents{subsection}[3.8em]{}%
    {\contentslabel{2.3em}}{\hspace*{-2.3em}}{\titlerule*[0.6em]{.}\contentspage}}
\vspace{1em}
\crefalias{section}{appendix}
\crefalias{subsection}{appendix}

\section{Model details}
\label{app:model}

\begin{figure}[t]
    \centering
    \begin{subfigure}[b]{0.40\linewidth}
      \centering
      \includegraphics[width=0.90\linewidth]{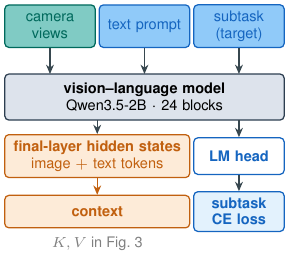}
      \caption{The VLM encoder}
      \label{fig:vlm-enc}
    \end{subfigure}\hfill
    \begin{subfigure}[b]{0.56\linewidth}
      \centering
      \includegraphics[scale=1.25]{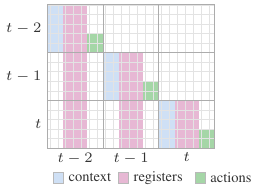}
      \caption{Attention pattern, $K\myeq3$}
      \label{fig:attn}
    \end{subfigure}
    \caption{(a) The VLM encoder.
    It feeds the expert of \cref{fig:expert} the encoded context.
    A vision--language model encodes the views and the prompt; its final-layer hidden states are the encoded context and its LM head predicts the subtask. (b) The registers and noisy action tokens of a frame cross-attend the context of that frame and self-attend the registers of the $K$ frames up to it, so the past reaches the current chunk only through the registers, and only these are kept for later frames.}
    \label{fig:vlm}
\end{figure}

\Cref{fig:vlm} shows how the prompt and the camera views are packed into one stream for the VLM of \cref{sec:context}.

\myparagraph{Prompt.} The prompt $z_t$ of \cref{sec:cond} is a structured user turn carrying one \texttt{field: value} per line:
\begin{verbatim}
embodiment: gr1
instruction: grasp the mug
speed: 2
fps: 20
control mode: absolute joint
\end{verbatim}
The speed field is the length of the episode in frames divided by 100 and rounded, so that a smaller value marks a shorter demonstration.
It is left out at evaluation.
Episodes of ABC~\citep{allshire2026abc} that are labelled with their outcome also carry a \texttt{success} field.

\myparagraph{Positions.} In the expert's self-attention, register $i$ of a frame sits at slot $i$ and action step $j$ at slot $\nreg+j$, so that the registers of each frame carry their own temporal encoding.
The frame and slot axes reuse the temporal and spatial axes of the VLM's multimodal rotary embedding.

\subsection{Inference cost and caching}
\label{app:latency}

\begin{figure}[t]
  \centering
  \includegraphics[width=\linewidth]{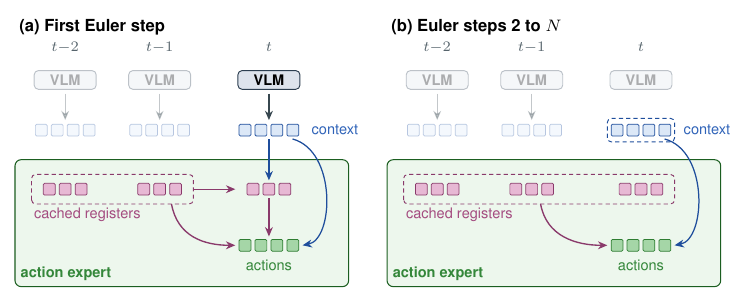}
  \caption{\textbf{Register cache at inference} ($K\myeq3$). (a) In the first Euler step of a chunk, the VLM encodes the current frame $t$. The registers of frame $t$ read its context and the cached registers of frames $t{-}1$ and $t{-}2$. The action tokens read the context and the registers of the three frames; past frames are not encoded again. (b) In the $N{-}1$ other Euler steps, only the action tokens are updated, and they read the context and the register keys and values from the cache. The registers do not read the action tokens and skip adaptive normalisation, so they do not depend on the flow time and are computed once per frame. When the next frame enters, the registers of frame $t{-}2$ are dropped.}
  \label{fig:register-cache}
\end{figure}

\myparagraph{What is cached.} One chunk is an encoder pass, then $N$ Euler steps over the action tokens, the first of which also computes the register tokens of the current frame (\cref{fig:register-cache}).
Only the noisy actions change from one Euler step to the next, so four quantities are reused across the $N$ steps.
\begin{enumerate}
    \setlength{\itemsep}{0pt}\setlength{\parskip}{0pt}\setlength{\topsep}{2pt}
    \item The encoded context $\ctx_t$ of \cref{eq:cond-context}.
    \item Its projection to keys and values, one pair per expert block.
    \item The keys and values of the register tokens $\mem_t$ of the current frame, one pair per expert block.
    \item The register keys and values of the $K-1$ preceding frames, carried over from earlier chunks.
\end{enumerate}
The first three are built the same way at training and at inference, while the fourth is not: the first registers of a window hold memory from earlier registers at inference, and none at training.

\myparagraph{What it costs.} \Cref{fig:latency-main} gives the measurement.
On one H100, a chunk takes 22.1\,ms without registers, 22.6\,ms at $K\myeq1$ and 22.8\,ms at $K\myeq3$, the window used for training and evaluation.
Even at $K\myeq32$, a chunk takes only 24.6\,ms.
The cache adds about 2\,MiB of GPU memory per past frame, 4\,MiB at $K\myeq3$, on a policy of 4.7\,GiB.

\myparagraph{Protocol.} Every measurement is taken on one H100, with two $256{\times}256$ cameras, a chunk of $10$ actions and $N\myeq10$ Euler steps, in bf16, with the expert and the VLM prefill run as CUDA graphs.
Each value is the median of two interleaved repetitions of 200 chunks.
The split by phase comes from a separate pass with a synchronisation after each phase, scaled to the median total.
The model is trained and evaluated at $K\myeq3$.
The longer windows are used only for these measurements.

\section{Data}
\label{app:data}

\subsection{Robot embodiments}
\label{app:embodiments}

\Cref{tab:embodiments} lists the robots of the pretraining mixture with their action conventions and source datasets.
From Open X-Embodiment, the mixture uses twelve datasets: Taco Play, Stanford Hydra, NYU Franka Play, BC-Z, RT-1, Furniture Bench, Austin Sailor, Austin BUDS, UT Austin MUTEX, Berkeley Cable Routing, VIOLA and Berkeley RPT.
The joint-delta actions of Berkeley RPT and RoboSet are converted to end-effector deltas by forward kinematics.
A robot appears under several conventions when its datasets command it differently, joint angles in one and end-effector deltas in another.
The state projection of \cref{sec:cond} is selected by state layout and control convention, and the action statistics of \cref{sec:data} are per dataset, so no source is remapped to another's convention.
A state layout is the ordered list of state channels a dataset records, for example the joint angles and gripper opening of each arm, and fixes the input width $d_s$ of its projection.

  \begin{table}[t]
  \centering
  \footnotesize
  \caption{\textbf{Robot embodiments in the training mixture.} ``EE'' = end-effector pose; ``J'' = joint angles; ``$\Delta$'' = delta (relative) commands, otherwise absolute; ``G'' = parallel gripper; ``DH'' = dexterous hand; ``W'' = waist/torso; ``B'' = mobile base.
  A robot may appear under several conventions across its source datasets.}
  \label{tab:embodiments}
  \begin{tabular}{ll>{\raggedright\arraybackslash}p{0.48\linewidth}}
  \toprule
  Robot & Action type & Sources \\
  \midrule
  \multicolumn{3}{l}{\emph{Single-arm}} \\
  WidowX & $\Delta$EE + G & \citep{walke2023bridgedata} \\
  Google Robot & $\Delta$EE + G & \citep{brohan2023rt1,jang2022bcz} \\
  Franka Panda & $\Delta$EE + G; J + G & \citep{oxe2024,bharadhwaj2024roboset,wu2025robomind,khazatsky2024droid,tian2026interndata} \\
  UR5e & J + G & \citep{wu2025robomind} \\
  \addlinespace
  \multicolumn{3}{l}{\emph{Dual-arm}} \\
  AgileX Cobot Magic & J + G & \citep{wu2025robocoin,wu2025robomind,liu2025rdt1b} \\
  ALOHA & J + G & \citep{tian2026interndata} \\
  AgiBot G1 & J + G & \citep{tian2026interndata} \\
  AgiBot G2A & J + G + W & \citep{agibotworld2026} \\
  Galaxea R1 Lite & J + G + W + B; J + G & \citep{jiang2025galaxea,wu2025robocoin} \\
  Dual Franka FR3 & J + G & \citep{wu2025robomind} \\
  AgiBot A2D & J + G & \citep{wu2025robocoin} \\
  Galbot G1 & J + G & \citep{wu2025robocoin} \\
  RealMan RMC-AIDA-L & J + G & \citep{wu2025robocoin} \\
  AI2 AlphaBot 2 & J + G & \citep{wu2025robocoin} \\
  YAM & J + G & \citep{allshire2026abc} \\
  Bimanual Panda & $\Delta$EE + G; $\Delta$EE + DH & \citep{nvidia2025gr00tn1} \\
  Fourier GR-1 & J + DH & \citep{nvidia2025gr00tn1} \\
  TienKung & J + G; J + DH & \citep{wu2025robomind} \\
  AIRBOT MMK2 & J + DH & \citep{wu2025robocoin} \\
  Unitree G1 & J + DH & \citep{wu2025robocoin} \\
  Kuavo & J + DH & \citep{wu2025robocoin} \\
  \addlinespace
  \multicolumn{3}{l}{\emph{Human}} \\
  Real Human & $\Delta$EE + eigengrasp & \citep{hoque2026egodex,punamiya2026egoverse,ropedia2026xperience10m} \\
  \bottomrule
  \end{tabular}
  \end{table}

\subsection{Data curation}
\label{app:curation}

Aggregating data across embodiments, simulators, and collection pipelines introduces heterogeneous noise: sensor outliers, temporal state-action misalignment, and inconsistent conventions across datasets sharing a robot.
\begin{itemize}
    \item \emph{Channel selection.} We keep only the degrees of freedom the policy commands, dropping derived or uncontrolled channels: forward-kinematic end-effector poses, joint velocities, efforts, IMU readings, and telepresence-parked head, waist, and leg joints.
    \item \emph{Degenerate episodes.} We discard static and near-zero-variance recordings, including episodes whose follower arm never moves, a failure mode Qwen-RobotManip reports for ${\sim}81\%$ of RoboMIND's UR subset.
    \item \emph{Sudden jumps.} We fit a smooth trend to each channel with median and Savitzky--Golay filters, and flag a frame when its residual and its acceleration or jerk exceed a per-channel threshold, the larger of 0.05 in raw units and four times the pooled 95th percentile.
    Grippers are exempt.
    Whole episodes are removed only from InternData-A1, where simulator collisions cause the jumps (6.3\% of its episodes).
    \item \emph{Action representation.} Rotations are harmonised to one Euler parameterisation; other conventions are kept as the source provides them.
    Each action dimension is mapped to $[-1,1]$ from its per-dataset 1st and 99th percentiles~\citep{wang2026qwenvla}, which removes scale differences across embodiments while preserving relative motion within each source.
    \item \emph{Convention correction.} We fix per-dataset sign conventions that would otherwise break cross-dataset consistency (\eg, a sign-flipped elbow joint in InternData); differing scales and encodings are then absorbed by the per-dataset normalisation.
\end{itemize}

\subsection{Sampling weights}
\label{app:sampling}

During pretraining, each dataset $d$ is sampled with probability proportional to $F_d^{0.5}\,L_d^{0.5}\,V_d^{0.35}$ (\cref{tab:hparams}). $F_d$ is its number of frames, and $L_d$ and $V_d$ are its language and vision diversity, each divided by its median over the datasets.
Datasets from the same source are grouped before weighting.

Both diversities are the effective rank of a set of L2-normalised embeddings, $2^{H(p)}$ with $p_i = \sigma_i^2 / \sum_j \sigma_j^2$, where $\sigma_i$ are the singular values of the embedding matrix.
The embeddings are not centred, so that a dataset with two distinct instructions does not collapse to rank one.
For the language diversity, the rows are the SigLIP~\citep{zhai2023siglip} text embeddings of the distinct episode task strings, each scaled by the square root of the fraction of frames it covers.
For the vision diversity, the rows are the DINOv3~\citep{simeoni2025dinov3} class tokens of every 15th frame of the main external camera, at most 50{,}000 frames per dataset.

The RoboCOIN and InternData-A1 collections are further weighted by 0.5, and the weights are capped so that no dataset is seen more than four times.

\subsection{Egocentric data}
\label{app:ego}

\myparagraph{Sources.} The three egocentric sources differ in capture conditions: EgoDex is seated tabletop recordings on Apple Vision Pro, EgoVerse head-mounted rigs, and Xperience in-the-wild moving-camera footage.
Frame rates range from 20 to 30\,Hz and are left unaltered; the differences are absorbed by the per-dataset quantile normalisation of \cref{sec:data}.

\myparagraph{Action target.} We follow the layout of Qwen-VLA's egocentric target~\citep{wang2026qwenvla}: per hand, a 6-dimensional wrist motion as translation and axis-angle rotation, and 10 articulation coefficients, 32 dimensions in total.
Unlike Qwen-VLA, which expresses the wrist relative to its own initial pose, we express it in the observation camera frame through the rigid transform $T_{\mathrm{cam}}^{-1} T_{\mathrm{wrist}}$.
The articulation coefficients are the first 10 principal components (``eigengrasps'') of the finger-joint configuration under a PCA basis shared across our three sources.
The target is defined relative to the frame that opens each action chunk, so it is composed at load time from the stored world-frame wrist poses and per-frame camera poses rather than precomputed.

\myparagraph{State and pacing.} The egocentric data have no proprioceptive state, and the policy receives a learned null state in its place.
Human hands move faster than a teleoperated robot, so the human actions are slowed down in time~\citep{yuan2026qwenrobot}.
Waypoint $i$ of a chunk is the hand pose at time $i\rho/f$ after the observation, where $f$ is the frame rate of the recording and $\rho \le 1$ a per-source pace (\cref{tab:hparams}).
The pose is interpolated between the two nearest frames.
A one-second chunk at $\rho = 0.25$ thus holds a quarter of a second of human motion.

\subsection{Vision--language data}
\label{app:vldata}

REKI is co-trained on vision--language data at 10\% of the training samples (\cref{tab:hparams}).
The pool is every FineVision~\citep{wiedmann2025finevision} dataset except the text-only ones, plus the LLaVA-OneVision-1.5~\citep{an2025llavaov15} datasets that FineVision does not already contain, ${\sim}20.2$M samples.
We keep the datasets covering the tasks most relevant to manipulation: natural-scene objects and attributes, spatial relations, counting, grounding, captioning, scene text, and knowledge and commonsense VQA.
Documents, charts, tables, mathematics, code, transcription, GUI grounding, memes and puzzles are dropped. 60 datasets remain, 7.64M samples or 37.8\% of the pool.
One question--answer pair is drawn per sample and sampling is uniform over samples, so exposure is proportional to the size of a dataset.

{\raggedright
The 51 kept from FineVision are \texttt{a\_okvqa}~\citep{schwenk2022aokvqa}, \texttt{alfworldgpt}~\citep{shridhar2021alfworld}, \texttt{allava\_laion}~\citep{chen2024allava}, \texttt{allava\_vflan}~\citep{chen2024allava}, \texttt{aokvqa}~\citep{schwenk2022aokvqa}, \texttt{art}~\citep{chng2019art}, \texttt{cambrian(filtered)\_processed}~\citep{tong2024cambrian}, \texttt{clevr}~\citep{lindstrom2022clevrmath}, \texttt{clevr\_math}~\citep{lindstrom2022clevrmath}, \texttt{coco\_colors}~\citep{karakus2024cococolors}, \texttt{cocoqa}~\citep{ren2015cocoqa}, \texttt{cocotext}~\citep{veit2016cocotext}, \texttt{densefusion\_1m}~\citep{li2024densefusion}, \texttt{drivelm}~\citep{sima2024drivelm}, \texttt{est\_vqa}~\citep{wang2020estvqa}, \texttt{google\_landmarks}~\citep{weyand2020landmarks}, \texttt{image\_textualization(filtered)}~\citep{pi2024imagetextualization}, \texttt{indoor\_qa}~\citep{quattoni2009indoor}, \texttt{laion\_gpt4v}~\citep{laion2023gpt4v}, \texttt{LLaVA\_Instruct\_150K}~\citep{liu2023llava}, \texttt{lnqa}~\citep{ponttuset2020narratives}, \texttt{localized\_narratives}~\citep{ponttuset2020narratives}, \texttt{lrv\_normal(filtered)}~\citep{liu2024lrv}, \texttt{lvis\_instruct4v}~\citep{wang2023lvisinstruct4v}, \texttt{mimic\_cgd}~\citep{laurencon2024whatmatters}, \texttt{mmevol}~\citep{luo2024mmevol}, \texttt{nlvr2}~\citep{suhr2017nlvr}, \texttt{objects365\_qa}~\citep{shao2019objects365}, \texttt{ocrvqa}~\citep{mishra2019ocrvqa}, \texttt{oodvqa}~\citep{tu2023unicorns}, \texttt{scienceqa}~\citep{lu2022scienceqa}, \texttt{scienceqa(nona\_context)}~\citep{lu2022scienceqa}, \texttt{sharegpt4o}~\citep{opengvlab2024sharegpt4o}, \texttt{sharegpt4v(coco)}~\citep{chen2024sharegpt4v}, \texttt{sharegpt4v(knowledge)}~\citep{chen2024sharegpt4v}, \texttt{sharegpt4v(llava)}~\citep{chen2024sharegpt4v}, \texttt{sharegpt4v(sam)}~\citep{chen2024sharegpt4v}, \texttt{sketchyvqa}~\citep{tu2023unicorns}, \texttt{spatialsense}~\citep{yang2019spatialsense}, \texttt{spot\_the\_diff}~\citep{jhamtani2018spotthediff}, \texttt{st\_vqa}~\citep{biten2019stvqa}, \texttt{super\_clevr(mathv360k)}~\citep{li2023superclevr}, \texttt{tallyqa}~\citep{acharya2019tallyqa}, \texttt{textcaps}~\citep{sidorov2020textcaps}, \texttt{textocr(gpt4v)}~\citep{carter2024textocrgpt4v}, \texttt{textvqa}~\citep{singh2019textvqa}, \texttt{vision\_flan(filtered)}~\citep{xu2024visionflan}, \texttt{visual7w}~\citep{zhu2016visual7w}, \texttt{vqav2}~\citep{goyal2017vqav2}, \texttt{vsr}~\citep{liu2023vsr}, and \texttt{wildvision}~\citep{lu2024wildvision}.
The nine kept from LLaVA-OneVision-1.5 are \texttt{gqa}~\citep{hudson2019gqa}, \texttt{llava\_cot\_100k}~\citep{xu2025llavacot}, \texttt{llava\_wild}~\citep{liu2023llava}, \texttt{llrv\_gpt4v}~\citep{liu2024lrv}, \texttt{sherlock}~\citep{hessel2022sherlock}, \texttt{svit}~\citep{zhao2023svit}, \texttt{viquae}~\citep{lerner2022viquae}, \texttt{vision\_oritented}, and \texttt{visual\_chat}.\par}

To these we add the object- and free-space-reference splits of RoboPoint~\citep{yuan2024robopoint}, the 2D and 3D splits of RefSpatial~\citep{zhou2025refspatial}, and PixMo-Points~\citep{deitke2024pixmo} rewritten as pointing and counting questions, all with coordinates on a $[0,1000]$ grid.

\section{Training configuration}
\label{app:training}

\Cref{tab:hparams} lists the training configuration.
\Cref{tab:setups} gives the node count, the number of steps and the settings that differ between pretraining and the five finetuning runs.
Settings absent from \cref{tab:setups} are those of \cref{tab:hparams} in every run.

\begin{table}[t]
\centering
\caption{\textbf{Training configuration.}}
\label{tab:hparams}
\small
\begin{tabular}{@{}l>{\raggedright\arraybackslash}p{0.59\linewidth}@{}}
\toprule
Setting & Value \\
\midrule
\multicolumn{2}{@{}l}{\textit{Encoder}}\\
Vision--language backbone      & Qwen3.5-2B (18 gated-DeltaNet + 6 full-attention layers) \\
Image resolution               & native, up to $448{\times}448$ \\
Expert conditioning            & cross-attention to the final-layer backbone states in every expert block \\
Trainable components           & backbone at a reduced LR (\cref{tab:setups}), last $4$ vision blocks, expert \\
Vision--language co-training   & $10\%$: FineVision (subset), LLaVA-OneVision-1.5 (subset), RoboPoint, RefSpatial, PixMo \\
Text loss (subtask, vision--language data) & cross-entropy, weight $0.1$, pretraining only \\
Prompt-field dropout           & $0.15$ (embodiment), $0.2$ (control frequency, speed) \\
\midrule
\multicolumn{2}{@{}l}{\textit{Action expert}}\\
Size                           & $10$ blocks, $768$ hidden, $12$ heads, $3072$ MLP \\
Precision                      & bfloat16, flow heads in fp32 \\
Proprioception                 & per-layout linear layer with GELU, shared zero-initialised projection, summed into the adaptive-normalisation conditioning \\
Proprioception dropout         & $0.2$ \\
Registers per frame            & $32$ \\
Window                         & $K = 3$ frames at a $1$\,s stride \\
\midrule
\multicolumn{2}{@{}l}{\textit{Action generation}}\\
Action chunk                   & $1$\,s ($\mathrm{round}(\text{control freq.})$ steps) \\
Objective                      & flow matching, DoF-balanced \\
Time-step distribution         & Beta \\
Flow draws per forward pass    & $8$ \\
\midrule
\multicolumn{2}{@{}l}{\textit{Data}}\\
Robot corpora                  & OXE, DROID, AgiBot, RoboMIND, Galaxea, RoboSet, RDT, RoboCOIN, GR00T, InternData, ABC \\
Egocentric human video         & EgoDex, EgoVerse, Xperience \\
Sampling weight                & $\propto \text{\{\#frames\}}^{0.5}\, \text{\{language diversity\}}^{0.5}\, \text{\{vision diversity\}}^{0.35}$ (\cref{app:sampling}) \\
Normalisation                  & quantile (1st / 99th percentile) \\
Action-speed alignment         & EgoDex $0.6$, EgoVerse $0.25$, Xperience $1$ (unchanged) \\
\midrule
\multicolumn{2}{@{}l}{\textit{Vision \& augmentation}}\\
Camera dropout                 & $0.15$ per camera in pretraining, $0.2$ in finetuning; at least one camera kept \\
Camera order                   & shuffled \\
Colour jitter                  & brightness, contrast, saturation $\times\,\mathcal{U}(0.8, 1.2)$, drawn for each frame of the window and shared by all cameras of the frame \\
Random crop                    & overhead and side cameras, scale $\mathcal{U}(0.95, 1)$, one box per camera for the window; pretraining and LIBERO only \\
\midrule
\multicolumn{2}{@{}l}{\textit{Optimisation}}\\
Learning rate                  & expert: $1{\times}10^{-4}$, VLM: $1{\times}10^{-5}$ to $3{\times}10^{-5}$ (\cref{tab:setups}) \\
Weight decay                   & $0$ \\
Gradient clipping              & $1.0$ \\
Schedule                       & constant with warmup ($1{,}000$ steps when finetuning the pretrained model, $5{,}000$ otherwise), then decay over the last $5{,}000$ steps (finetuning only) \\
Weight averaging               & EMA, decay $0.999$ \\
Gradient checkpointing         & \cmark \\
\bottomrule
\end{tabular}
\end{table}

\begin{table}[ht]
  \centering
  \caption{\textbf{Per-run setup of REKI} for pretraining and the five finetuning runs, and the number of evaluation rollouts per task.
  The two RoboTwin runs are finetuned on the Clean demonstrations alone (Clean2Rand) and on Clean and Randomised together (in-distribution).
  $^\ast$Multiplied by the language and vision diversity terms of \cref{app:sampling}.}
  \label{tab:setups}
  \small
\adjustbox{max width=\linewidth}{%
\begin{tabular}{lcccccc}
  \toprule
   & \multirow{2}{*}{Pretraining} & \multirow{2}{*}{LIBERO} & \multicolumn{2}{c}{RoboTwin} & \multirow{2}{*}{RoboCasa365} & \multirow{2}{*}{RoboCasa-GR1} \\
  \cmidrule(lr){4-5}
   & & & Clean & Clean+Rand & & \\
  \midrule
  Nodes ($8\times$H100)                   & 4      & 1      & 1      & 3      & 4 & 3 \\
  Steps                                   & 200k   & 15k    & 20k    & 25k    & 25k & 25k \\
  Wall-clock time (h)                     & 176    & 5.2    & 10.9   & 14.2   & 12.5 & 8.3 \\
  GPU-hours                               & 5{,}632 & 41    & 87     & 341    & 401 & 199 \\
  Backbone LR ($\times$ expert LR)        & 0.1    & 0.3    & 0.2    & 0.3    & 0.3 & 0.2 \\
  Sampling weights $\propto\sqrt{\#\text{frames}}$ & \cmark$^\ast$ & \xmark & \xmark & \cmark & \xmark & \xmark \\
  Overhead-camera dropout                 & \xmark  & \cmark   & \xmark  & \xmark  & \xmark & \xmark \\
  Mirroring (\cref{sec:augmentation}) &  \xmark      & \xmark    & \cmark    & \cmark    & \cmark & \cmark \\
  Evaluation rollouts per task            & --     & 50     & 20     & 20     & 50 & 50 \\
  \bottomrule
  \end{tabular}
}
  \end{table}

\subsection{Camera selection}
\label{app:cameras}

In pretraining, two rules remove cameras from a sample after the random camera dropout.
When two or more side cameras remain, one of them is kept at random and the others are removed.
On a two-arm robot with both wrist cameras present, the wrist camera of a static arm is removed.
An arm is static when, over the action chunk, every action dimension of the arm varies by less than $0.1$ of its standard deviation in the dataset.
Both rules keep at least one camera other than the overhead camera.
Finetuning uses neither rule.
Random camera dropout never removes the overhead camera, except in LIBERO finetuning (\cref{tab:setups}).

\subsection{Mirroring}
\label{app:mirroring}

A sample is mirrored with probability $0.5$ in the finetuning runs on RoboTwin, RoboCasa365 and RoboCasa-GR1 (\cref{tab:setups}).
The images are flipped horizontally, and the state and the actions are reflected through the same plane.
On a two-arm robot, the two arms and their wrist cameras are swapped.
The prompt of a mirrored sample starts with \texttt{[MIRROR WORLD]}, and its state goes through a separate projection with separate normalisation statistics.
The policy is evaluated in the unmirrored world only.

\section{Full benchmark results}
\label{app:results}

\Cref{tab:libero-plus,tab:robotwin-c2r} give the per-condition results summarised in \cref{fig:ood-size}, together with the in-distribution results on LIBERO and RoboTwin.
Each LIBERO-Plus column perturbs one dimension: the camera viewpoint, the initial robot state, the language instruction, the lighting, the background texture, the sensor noise or the object layout.
The Overall column is the success rate over all 10,030 perturbed episodes, so each dimension is weighted by its number of episodes~\citep{fei2025liberoplus}.
For the in-distribution RoboTwin results, the models are finetuned on clean and randomised scenes together and evaluated on each (Clean, Rand).
For Clean2Rand, the models are finetuned on clean scenes only and evaluated on clean scenes (Easy), under a randomised background, lighting, clutter or table height applied alone, and under all four together (Hard).
The baseline numbers are taken from published results, and \cref{tab:provenance} gives the source of each one.
The Clean2Rand row of Qwen-RobotManip~\citep{yuan2026qwenrobot} is its joint-space variant.
The LIBERO-Plus results of $\pi_0$ and $\pi_{0.5}$ are from the zero-shot block of ACoT-VLA~\citep{zhong2026acotvla}, which evaluated the released checkpoints.
ACoT-VLA is initialised from $\pi_{0.5}$, so it is grouped with the models pretrained on in-house data.
The LIBERO-Plus results of StarVLA are those of its Qwen3-VL-OFT model in the StarVLA repository.
For Being-H0.7 and ABot-M0.5, the LIBERO-Plus Overall is pooled from the seven published dimensions, because the published total does not match them.
The RoboCasa365 leaderboard is at \url{https://robocasa.ai/leaderboard}.
On RoboCasa-GR1, the result of GR00T-N1.5 is that of the released checkpoint in the Isaac-GR00T repository (\url{https://github.com/NVIDIA/Isaac-GR00T}, branch \texttt{n1.5-release}), and the result of $\pi_{0.5}$ is the one reported by StarVLA, which states no source for it.
Parameter counts include the action expert, are rounded from exact counts and are read off the released checkpoints where available, including those of StarVLA and RLDX-1~\citep{kim2026rldx1}.
Qwen-RobotManip states no expert size; our expert follows its design, so we count 0.2B for it.

\begin{table}[t]
\centering
\caption{\textbf{LIBERO and LIBERO-Plus results.} All models are finetuned on LIBERO; LIBERO-Plus evaluates out-of-distribution robustness across seven perturbation dimensions.
\emph{Italic}: model not released at the time of writing. Bold: best in each column within each block, italic rows excluded. $^\dagger$Pooled from the seven reported dimensions, which the published total does not match. }
\label{tab:libero-plus}
\adjustbox{max width=\linewidth}{%
\begin{tabular}{lcccccccccccc}
\toprule
\multirow{2}{*}{Method} & \multicolumn{2}{c}{\#Params} & \multirow{2}{*}{LIBERO} & \multicolumn{8}{c}{LIBERO-Plus} \\
\cmidrule(lr){2-3} \cmidrule(lr){5-12}
& VLM & Expert & & Camera & Robot & Language & Light & Backgd & Noise & Layout & Overall \\
\midrule
\multicolumn{5}{l}{\emph{No pretraining}} \\
StarVLA~\citep{starvla2026codebase} & 4.4B & 0.1B  & 98.0 & 47.0 & 60.1 & 87.0 & 96.3 & 95.3 & 73.1 & \textbf{79.2} & 75.0 \\
Qwen-RobotManip-scratch~\citep{yuan2026qwenrobot} & 4.5B & 0.2B  & \textbf{98.2} & 70.4 & 44.9 & \textbf{88.1} & 95.8 & 95.5 & 84.4 & 79.1 & 78.3 \\
\textbf{REKI (ours)}  \quad\emph{scratch} & 2.2B & 0.2B & 96.5 & \textbf{94.4} & \textbf{63.9} & 80.7 & \textbf{97.5} & \textbf{96.0} & \textbf{96.0} & \textbf{79.2} & \textbf{86.1} \smallskip \\
\midrule
\multicolumn{5}{l}{\emph{With pretraining -- in-house data}} \\
$\pi_0$~\citep{black2024pi0}                 & 2.9B & 0.3B  & 94.4 & 61.0 & 40.8 & 63.5 & 89.3 & 84.1 & 80.1 & 76.4 & 69.4 \\
$\pi_{0.5}$~\citep{intelligence2025pi05}             & 2.9B & 0.4B  & 97.6 & 75.8 & 79.4 & 83.3 & 95.5 & 95.0 & 89.6 & 87.0 & 85.7 \\
ACoT-VLA~\citep{zhong2026acotvla} & 2.9B & 0.9B  & 98.5 & 72.6 & 82.6 & 87.5 & \textbf{97.7} & 96.5 & 87.8 & \textbf{88.1} & 86.6 \\
\nr{Being-H0.7}~\citep{luo2026being} & \nr{3B} & -- & \nr{99.2} & \nr{82.0} & \nr{59.0} & \nr{82.8} & \nr{97.8} & \nr{90.0} & \nr{93.5} & \nr{88.5} & \nr{84.0}$^\dagger$ \\
\nr{Qwen-RobotManip}~\citep{yuan2026qwenrobot} & \nr{4.5B} & \nr{0.2B} & \nr{99.1} & \nr{87.2} & \nr{75.5} & \nr{85.6} & \nr{96.6} & \nr{97.7} & \nr{97.7} & \nr{87.3} & \nr{89.0} \\
RLDX-1~\citep{kim2026rldx1} & 5.3B & 1.6B & 97.8 & 75.0 & \textbf{91.8} & \textbf{92.2} & 97.1 & 95.5 & 80.3 & 80.7 & 86.7 \\
\addlinespace
\multicolumn{5}{l}{\emph{With pretraining -- public data only}} \\
ABot-M0~\citep{yang2026abot} & 4.4B & 0.2B & 98.6 & 60.4 & 67.9 & 86.4 & 96.2 & 91.6 & 86.4 & 82.6 & 80.5 \\
ABot-M0.5~\citep{chen2026abotm05} & 5B & -- & \textbf{99.4} & 70.5 & 87.4 & 88.6 & 94.0 & 89.7 & 75.5 & 85.2 & 83.7$^\dagger$ \\
InternVLA-A1.5~\citep{ma2026internvla} & 2.2B & 0.5B & 98.9 & 83.1 & 55.1 & 86.9 & 96.4 & 98.2 & 95.6 & 85.2 & 84.8 \\
\textbf{REKI (ours)} & 2.2B & 0.2B & 97.5 & \textbf{93.2} & 75.6 & 88.5 & \textbf{97.7} & \textbf{98.4} & \textbf{98.6} & 84.7 & \textbf{90.4} \\
\bottomrule
\end{tabular}
}
\vspace{-0.1cm}
\end{table}

  \begin{table}[t]
  \centering
  \footnotesize
  \caption{\textbf{RoboTwin results.} \textit{Left}: in-distribution results for models finetuned on clean and randomised scenes.
  \textit{Right}: out-of-distribution evaluation on RoboTwin Clean2Rand, where models are finetuned on the Clean dataset only and tested under various environment randomisations.
  \emph{Italic}: model not released at the time of writing. Bold: best in each column within each block, italic rows excluded.}
  \label{tab:robotwin-c2r}
\adjustbox{max width=\linewidth}{%
\begin{tabular}{lccccccccccc}
  \toprule
  \multirow{2}{*}{Method} & \multicolumn{2}{c}{\#Params} & \multicolumn{2}{c}{RoboTwin} && \multicolumn{6}{c}{RoboTwin Clean2Rand} \\
  \cmidrule(lr){2-3} \cmidrule(lr){4-5} \cmidrule(lr){7-12}
   & VLM & Expert & Clean & Rand && Easy & Backgd & Light & Clutter & Height & Hard \\
  \midrule
  \multicolumn{5}{l}{\emph{No pretraining}} \\
  StarVLA~\citep{starvla2026codebase} & 4.4B & 0.1B & 88.2 & 88.3 && 58.1 & 27.1 & 50.9 & 24.2 & 48.4 & 10.6 \\
  Qwen-RobotManip-scratch~\citep{yuan2026qwenrobot} & 4.5B & 0.2B & \textbf{88.7} & \textbf{88.4} && \textbf{71.6} & 60.6 & \textbf{70.7} & 24.6 & \textbf{63.6} & 22.6 \\
    \textbf{REKI (ours)} \quad \emph{scratch} & 2.2B & 0.2B & 83.0 & 83.8 & & 67.1 & \textbf{62.4} & 68.2 & \textbf{27.5} & 60.7 & \textbf{26.3}  \smallskip \\
  \midrule
      \multicolumn{5}{l}{\emph{With pretraining -- in-house data}} \\
  $\pi_0$~\citep{black2024pi0} & 2.9B & 0.3B & 65.9 & 58.4 && -- & -- & -- & -- & -- & -- \\
  $\pi_{0.5}$~\citep{intelligence2025pi05} & 2.9B & 0.4B & 82.7 & 76.8 && 73.1 & 67.0 & 69.2 & \textbf{57.9} & 67.6 & 47.9 \\
  GR00T-N1.7~\citep{nvidia2025gr00tn1} & 1.5B & 1.6B & -- & -- && 43.6 & 40.4 & 41.9 & 27.1 & 39.0 & 20.7 \\
  \nr{Being-H0.7}~\citep{luo2026being} & \nr{3B} & -- & \nr{90.2} & \nr{89.6} && -- & -- & -- & -- & -- & -- \\
  \nr{Qwen-RobotManip}~\citep{yuan2026qwenrobot} & \nr{4.5B} & \nr{0.2B} & \nr{93.4} & \nr{92.5} && \nr{73.2} & \nr{74.6} & \nr{68.4} & \nr{61.3} & \nr{71.0} & \nr{62.6} \\
  \addlinespace
          \multicolumn{5}{l}{\emph{With pretraining -- public data only}} \\
  ABot-M0~\citep{yang2026abot} & 4.4B & 0.2B & 86.1 & 85.1 && 70.7 & 56.5 & 68.8 & 46.0 & 56.3 & 36.0 \\
  ABot-M0.5~\citep{chen2026abotm05} & 5B & -- & \textbf{94.0} & \textbf{94.2} && -- & -- & -- & -- & -- & -- \\
  InternVLA-A1.5~\citep{ma2026internvla} & 2.2B & 0.5B & 93.3 & 93.0 && -- & -- & -- & -- & -- & -- \\
  \textbf{REKI (ours)} & 2.2B & 0.2B & 85.9 & 84.6 && \textbf{73.4} & \textbf{71.3} & \textbf{73.5} & 57.8 & \textbf{69.9} & \textbf{48.7} \smallskip \\
  \bottomrule
  \end{tabular}
}
  \end{table}

\begin{table}[ht]
\centering
\footnotesize
\caption{\textbf{RoboCasa365 results} behind \cref{fig:robocasa}, on the 50 target tasks of the multi-task setting.
Hatched in \cref{fig:robocasa}, and marked $\ast$ here: pretrained on in-house data.
\emph{Italic}: model not released at the time of writing.}
\label{tab:robocasa365}
\begin{tabular}{lccccc}
\toprule
\multirow{2}{*}{Method} & \multirow{2}{*}{\#Params} & \multirow{2}{*}{Atomic} & \multicolumn{2}{c}{Composite} & \multirow{2}{*}{Avg} \\
\cmidrule(lr){4-5}
 & & & Seen & Unseen & \\
\midrule
Xiaomi-Robotics-1$^\ast$~\citep{xiaomi2026robotics1}   & 5.1B & 80.2 & 57.1 & 32.1 & 57.4 \\
ABot-M0.5~\citep{chen2026abotm05}                        & 5B   & 75.9 & 38.3 & 2.7  & 40.4 \\
RLDX-1$^\ast$~\citep{kim2026rldx1}                       & 6.9B & 67.6 & 27.9 & 8.5  & 36.0 \\
\nr{Qwen-RobotManip}$^\ast$~\citep{yuan2026qwenrobot}    & \nr{4.7B} & \nr{68.6} & \nr{20.1} & \nr{14.9} & \nr{35.9} \\
GR00T-N1.5$^\ast$~\citep{nvidia2025gr00tn1}              & 2.4B & 50.7 & 14.8 & 2.7  & 23.9 \\
GR00T-N1.6$^\ast$~\citep{nvidia2025gr00tn1}              & 3.0B & 51.1 & 9.4  & 1.7  & 21.9 \\
$\pi_{0.5}$$^\ast$~\citep{intelligence2025pi05}            & 3.3B & 39.6 & 7.1  & 1.2  & 16.9 \\
\midrule
\textbf{REKI (ours)}                                     & 2.4B & 65.8 & 32.0 & 14.1 & 38.4 \\
\bottomrule
\end{tabular}
\end{table}

\begin{table}[ht]
\centering
\footnotesize
\caption{\textbf{Source of each baseline result} in \cref{tab:libero-plus,tab:robotwin-c2r,tab:robocasa365} and \cref{fig:gr1}.
Paper: the paper of the model.
The name of another model: the paper of that model, which reports the result.
Checkpoint card: the card of the released checkpoint.
Repository: the per-task table of the released checkpoint in the README of the repository of the model.
Leaderboard: the RoboCasa365 leaderboard.
--: no published result.}
\label{tab:provenance}
\adjustbox{max width=\linewidth}{%
\begin{tabular}{lllllll}
\toprule
Method & LIBERO & LIBERO-Plus & RoboTwin Clean, Rand & RoboTwin Clean2Rand & RoboCasa365 & RoboCasa-GR1 \\
\midrule
\multicolumn{7}{l}{\emph{No pretraining}} \\
StarVLA & Qwen-RobotManip & repository & checkpoint card & Qwen-RobotManip & -- & paper \\
Qwen-RobotManip-scratch & paper & paper & paper & paper & -- & -- \\
\midrule
\multicolumn{7}{l}{\emph{With pretraining -- in-house data}} \\
$\pi_0$ & Qwen-RobotManip & ACoT-VLA & InternVLA-A1.5 & -- & -- & -- \\
$\pi_{0.5}$ & Qwen-RobotManip & ACoT-VLA & Qwen-RobotManip & Qwen-RobotManip & leaderboard & StarVLA \\
ACoT-VLA & paper & paper & -- & -- & -- & -- \\
GR00T-N1.5 & -- & -- & -- & -- & leaderboard & repository \\
GR00T-N1.6 & -- & -- & -- & -- & leaderboard & -- \\
GR00T-N1.7 & -- & -- & -- & Qwen-RobotManip & -- & -- \\
Being-H0.7 & Qwen-RobotManip & Qwen-RobotManip & paper & -- & -- & paper \\
Qwen-RobotManip & paper & paper & paper & paper & paper & -- \\
RLDX-1 & paper & paper & -- & -- & leaderboard & paper \\
Xiaomi-Robotics-1 & -- & -- & -- & -- & leaderboard & -- \\
\addlinespace
\multicolumn{7}{l}{\emph{With pretraining -- public data only}} \\
ABot-M0 & paper & paper & paper & Qwen-RobotManip & -- & paper \\
ABot-M0.5 & paper & paper & paper & -- & paper & -- \\
InternVLA-A1.5 & paper & paper & paper & -- & -- & -- \\
\bottomrule
\end{tabular}
}
\end{table}

\subsection{Pretraining steps}
\label{app:pretraining-steps}

\begin{figure}[ht]
  \centering
  \includegraphics{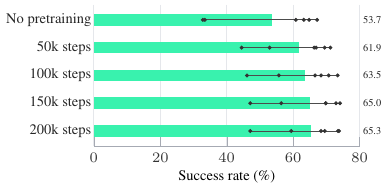}
  \caption{\textbf{Pretraining.} RoboTwin Clean2Rand success rate of REKI against the number of pretraining steps.
  The bars give the mean over the six conditions, and the dots the individual conditions.}
  \label{fig:pretrain-scaling}
\end{figure}

\Cref{fig:pretrain-scaling} gives the Clean2Rand success of REKI after 50k to 200k pretraining steps, and without pretraining.
All models are finetuned on clean RoboTwin scenes as in \cref{tab:robotwin-c2r}, but with a backbone learning rate of 0.1 instead of 0.2 times the expert learning rate, and without learning-rate decay.
The mean success rises from 53.7 without pretraining to 61.9 after 50k steps, and then by 3.4 points more up to 200k steps.
The first 50k steps add most under clutter (19.5 points) and under the Hard condition (11.7 points).

\subsection{Finetuning ablations}
\label{app:ablations}

\Cref{tab:ablations} compares the LIBERO finetuning configuration of REKI with variants that add mirroring, remove one augmentation or lower the backbone learning rate, all evaluated on LIBERO-Plus.
It also compares finetuning on RoboCasa365 with and without mirroring.
All variants use a memory window of $K\myeq3$ and the same number of finetuning steps.

\myparagraph{Mirroring.} On RoboCasa365, mirroring (\cref{app:mirroring}) adds 4.9 points on average, mostly on the atomic tasks ($+6.2$) and the seen composite tasks ($+7.9$).
On LIBERO-Plus, mirroring improves four of the seven perturbation dimensions, by $0.2$ to $1.7$ points.
However, it lowers the camera dimension by $6.5$ points, so the overall score drops by $0.6$.
A possible explanation is that the policy infers whether the world is mirrored from the images as well as from the prompt.
A large change of the camera viewpoint may then confuse the policy about whether the world is mirrored.

\myparagraph{Visual augmentation.} Disabling camera dropout or the image augmentations (colour jittering and random cropping) causes a large drop on the camera dimension of LIBERO-Plus, by $21.5$ and $23.6$ points.
This suggests that these augmentations make the policy robust to changes of the camera viewpoint.
They also give a small gain on the sensor noise and object layout dimensions, of $1.6$ to $4.6$ points.
For camera dropout, a likely reason is the wrist camera: the policy is trained to act from the wrist image alone, and the camera and sensor-noise perturbations of LIBERO-Plus change only the image of the external camera~\citep{fei2025liberoplus}.

\myparagraph{State dropout.} Disabling state dropout lowers the robot dimension by $8.1$ points: state dropout makes the policy robust to changes of the initial state of the robot.

\myparagraph{Backbone learning rate.} Lowering the backbone learning rate from $0.3$ to $0.1$ times the expert learning rate lowers the overall score by $3.3$ points.
The largest drops are on the camera and robot dimensions, by $11.9$ and $4.6$ points.

\begin{table}[t]
\centering
\caption{\textbf{Finetuning ablations}, memory window $K\myeq3$.
Each row changes one setting of the reported configuration (bold).}
\label{tab:ablations}
\small
\adjustbox{max width=\linewidth}{%
\begin{tabular}{@{}lcccccccc@{}}
\toprule
LIBERO-Plus & Camera & Robot & Language & Light & Backgd & Noise & Layout & Overall \\
\midrule
\textbf{REKI}                        & 93.2 & 75.6 & 88.5 & 97.7 & 98.4 & 98.6 & 84.7 & 90.4 \\
With mirroring                       & 86.7 & 77.3 & 88.7 & 98.6 & 99.6 & 98.3 & 84.3 & 89.8 \\
Without colour jitter and crop       & 69.6 & 74.8 & 90.4 & 97.9 & 97.8 & 96.6 & 80.1 & 85.7 \\
Without camera dropout               & 71.7 & 76.5 & 90.8 & 99.0 & 98.4 & 94.7 & 83.1 & 86.8 \\
Without state dropout                & 95.4 & 67.5 & 87.1 & 98.4 & 97.8 & 98.3 & 82.8 & 89.0 \\
Backbone LR $0.1\times$ expert LR    & 81.3 & 71.0 & 86.7 & 98.5 & 98.0 & 97.8 & 82.8 & 87.1 \\
\bottomrule
\end{tabular}}

\medskip
\begin{tabular}{@{}lcccc@{}}
\toprule
\multirow{2}{*}{RoboCasa365} & \multirow{2}{*}{Atomic} & \multicolumn{2}{c}{Composite} & \multirow{2}{*}{Avg} \\
\cmidrule(lr){3-4}
 & & Seen & Unseen & \\
\midrule
Without mirroring       & 59.6 & 24.1 & 13.5 & 33.5 \\
\textbf{With mirroring} & 65.8 & 32.0 & 14.1 & 38.4 \\
\bottomrule
\end{tabular}
\end{table}

\subsection{Training budgets of the compared methods}
\label{app:budgets}

\Cref{tab:training-budgets-samples} collects the training budgets published for the methods compared in \cref{tab:libero-plus,tab:robotwin-c2r}.
Qwen-RobotManip, its scratch variant, ABot-M0.5 and GR00T-N1.7 state no budget for the compared runs and are left out, as is ACoT-VLA, which is built on $\pi_{0.5}$.
No compared method reports wall-clock time or accelerator-hours, so the success rates of those tables are not matched for training compute.

\begin{table}[t]
\centering
\footnotesize
\setlength{\tabcolsep}{3.5pt}
\caption{\textbf{Training budgets} of the compared methods as optimiser steps, global batch size and training samples seen (batch $\times$ steps).
A sample is one timestep with the action chunk that follows it.
None of the methods reports accumulating gradients.
--: not reported or not applicable.
Values not stated in the papers are read off the released code or checkpoints.
Being-H0.7 states an approximate batch of 128 packed trajectory chunks for all its finetuning runs.
For REKI, the batch is $K\myeq3$ frames times the number of windows.
This is an upper bound, since a batch is trimmed to a token budget.
A slash separates the Clean and Clean+Rand RoboTwin runs.}
\label{tab:training-budgets-samples}
\adjustbox{max width=\linewidth}{%
\begin{tabular}{l ccc ccc ccc}
\toprule
\multirow{2}{*}{Method}
 & \multicolumn{3}{c}{Pretraining}
 & \multicolumn{3}{c}{LIBERO}
 & \multicolumn{3}{c}{RoboTwin} \\
\cmidrule(lr){2-4} \cmidrule(lr){5-7} \cmidrule(lr){8-10}
 & Steps & Batch & Samples
 & Steps & Batch & Samples
 & Steps & Batch & Samples \\
\midrule
\multicolumn{10}{l}{\emph{No pretraining}} \\
StarVLA~\citep{starvla2026codebase}
 & -- & -- & --
 & -- & -- & --
 & 140k & 192 & 26.9M
 \\
\textbf{REKI (ours)}, scratch
 & -- & -- & --
 & 30k & 720 & 21.6M
 & 40k / 45k & 720 / 2160 & 28.8M / 97.2M
 \\
\midrule
\multicolumn{10}{l}{\emph{With pretraining -- in-house data}} \\
$\pi_0$~\citep{black2024pi0}
 & 700k & -- & --
 & 30k & 32 & 0.96M
 & -- & -- & -- \\
$\pi_{0.5}$~\citep{intelligence2025pi05}
 & 280k\,+\,80k & -- & --
 & 30k & 256 & 7.7M
 & -- & -- & -- \\
Being-H0.7~\citep{luo2026being}
 & -- & -- & --
 & -- & 128 & --
 & -- & 128 & -- \\
RLDX-1~\citep{kim2026rldx1}
 & 100k & 8192 & 819.2M
 & 60k & 256 & 15.4M
 & -- & -- & -- \\
\addlinespace
\multicolumn{10}{l}{\emph{With pretraining -- public data only}} \\
ABot-M0~\citep{yang2026abot}
 & 100k & 1024 & 102.4M
 & 40k & 64 & 2.6M
 & 125k & 192 & 24.0M \\
InternVLA-A1.5~\citep{ma2026internvla}
 & 300k\,+\,600k & 1024 & 921.6M
 & 100k & -- & --
 & 100k & 384 & 38.4M \\
\textbf{REKI (ours)} (\cref{tab:setups})
 & 200k & 2880 & 576.0M
 & 15k & 720 & 10.8M
 & 20k / 25k & 720 / 2160 & 14.4M / 54.0M \\
\bottomrule
\end{tabular}
}
\end{table}

\subsection{Inference latency of the compared methods}
\label{app:latency-runs}

In \cref{sec:latency}, each model is timed with its released checkpoint and inference code on one H100 (driver 570, CUDA 12.8), one model per process, at batch size one.
The observations are synthetic and seeded, with new images and state at every chunk.
In the LIBERO setting, each model receives two $256{\times}256$ images, an 8-D state and a fixed prompt, and applies its own resizing and tokenisation.
Each repetition is one warm-up episode followed by 320 timed chunks, and runs as a separate job on its own GPU of a node reserved for the timing.
ABot-M0.5 takes several seconds per chunk, so each of its repetitions times 80 chunks, in four episodes of 20.
The end-to-end time runs from the numpy observation to the unnormalised numpy actions.
The model time starts from preprocessed tensors on the device, and the GPU time sums the CUDA kernel durations of a chunk, from a profiler pass taken last in the process.
Eager runs use inference mode where the released code does, and fp32 runs enable TF32, as the LeRobot evaluation~\citep{cadene2024lerobot} does.
Every run has to load the released weights strictly, return finite actions of the right shape that change with the input, and keep at least 95\% of them within the action range of the checkpoint.
The official compiled path of InternVLA-A1.5 is slower than eager, and its \texttt{optimized} path returns non-finite actions after the first chunk, which we avoid by keeping one autocast context open for the whole process.
For X-VLA, ABot-M0, InternVLA-A1.5 and ABot-M0.5, we also build a compiled path over the released code, without changing the computation.
These runs show what the models can reach, not what their users get.
The phase times of REKI and InternVLA-A1.5 in \cref{sec:latency} come from a separate run on one exclusive node (\cref{tab:latency-phases}).

\begin{table}[t]
  \centering
  \caption{\textbf{Phase times} of one chunk with LIBERO inputs, in ms of wall time, from a separate run. REKI runs the vision encoder and the VLM prefill as one CUDA graph, timed together.}
  \label{tab:latency-phases}
  {\footnotesize
  \begin{tabular}{@{}lrrr@{}}
  \toprule
  & REKI & \multicolumn{2}{c}{InternVLA-A1.5} \\
  \cmidrule(l){3-4}
  Phase & compiled & \texttt{optimized} & compiled by us \\
  \midrule
  Vision encoder & \multirow{2}{*}{9.8} & 11.5 & 2.1 \\
  VLM prefill & & 28.7 & 4.7 \\
  Euler steps (10) & 8.2 & 46.0 & 34.9 \\
  Other & 2.0 & 2.9 & 2.7 \\
  \midrule
  Model time & 19.7 & 88.2 & 44.1 \\
  \bottomrule
  \end{tabular}}
\end{table}

\section{Real robot experiments}
\label{app:real-robot}

\subsection{Tasks and scoring}

We evaluate on five tasks.
Partial credit is awarded for each completed subtask, such as a successful grasp or placement.
For tasks with an imposed ordering, a penalty is applied when objects are handled in the wrong order.
An episode counts as a full success only when every subtask is completed in the correct order.
The first two tasks involve a single object and are shorter-horizon.
The remaining three require manipulating two or three objects in a prescribed order, so the policy must keep track of which steps have already been completed.

\begin{itemize}
    \item \textbf{place\_banana\_in\_box}: pick the yellow banana and place it in the box.
    Partial credit: 0.5 for grasp; 0.5 for placement (only 0.25 if partially inside the box).
    \item \textbf{sweep\_paper\_into\_dustpan}: grasp the brush and sweep the paper crumples into the dustpan.
    Partial credit: 0.5 for grasp; 0.25 once one crumple is inside the dustpan; 0.25 once all crumples are inside.
    \item \textbf{arrange\_3\_fruits\_in\_plates}: place the purple grapes in the yellow plate, then the yellow banana in the pink plate, then the red strawberry in the blue plate.
    Partial credit: $1/6$ per grasp and $1/6$ per placement for each of the three fruits; $-1/6$ for wrong order.
    \item \textbf{stack\_3\_cubes}: stack the pink cube on the yellow cube, then the green cube on the pink cube.
    Partial credit: 0.25 per grasp and 0.25 per placement for each of the two cubes; $-0.25$ for wrong order.
    \item \textbf{stack\_3\_cups}: stack the pink cup into the yellow cup, then stack the blue cup into the pink cup.
    Partial credit: 0.25 per grasp and 0.25 per placement for each of the two cups; $-0.25$ for wrong order.
\end{itemize}

\subsection{Finetuning details}
\begin{table}[ht]
\centering
\small
\caption{\textbf{Finetuning hyperparameters} for the real-robot experiments.}
\label{tab:real-robot-hparams}
\begin{tabular}{@{}lcc@{}}
\toprule
 & REKI & $\pi_{0.5}$ \\
\midrule
Optimiser steps & 3{,}000 & 3{,}000 \\
Global batch size & 360 & 360 \\
Learning rate (action expert) & $5\times10^{-5}$ & $5\times10^{-5}$ \\
Learning rate (backbone) & $5\times10^{-6}$ & $5\times10^{-6}$ \\
LR schedule & warmup, then constant & warmup, then constant \\
Trainable vision blocks & last 4 & last 4 \\
Action chunk size $H$ & 30 (2\,s) & 50 (3.3\,s) \\
Memory window $K$ & $K\myeq3$ frames, 2\,s apart ($K\myeq1$ without memory) & -- \\
H100 GPU-hours & 8 & -- \\
\bottomrule
\end{tabular}
\end{table}

Both REKI and $\pi_{0.5}$ are initialised from their pretrained checkpoints and finetuned jointly on the five tasks, using the hyperparameters in \cref{tab:real-robot-hparams}.
The action chunk has $H\myeq30$ steps ($2$\,s) for REKI and $H\myeq50$ ($3.3$\,s) for $\pi_{0.5}$, which is fixed by its pretraining.

\begin{figure}[t]
\centering
\includegraphics[width=\linewidth]{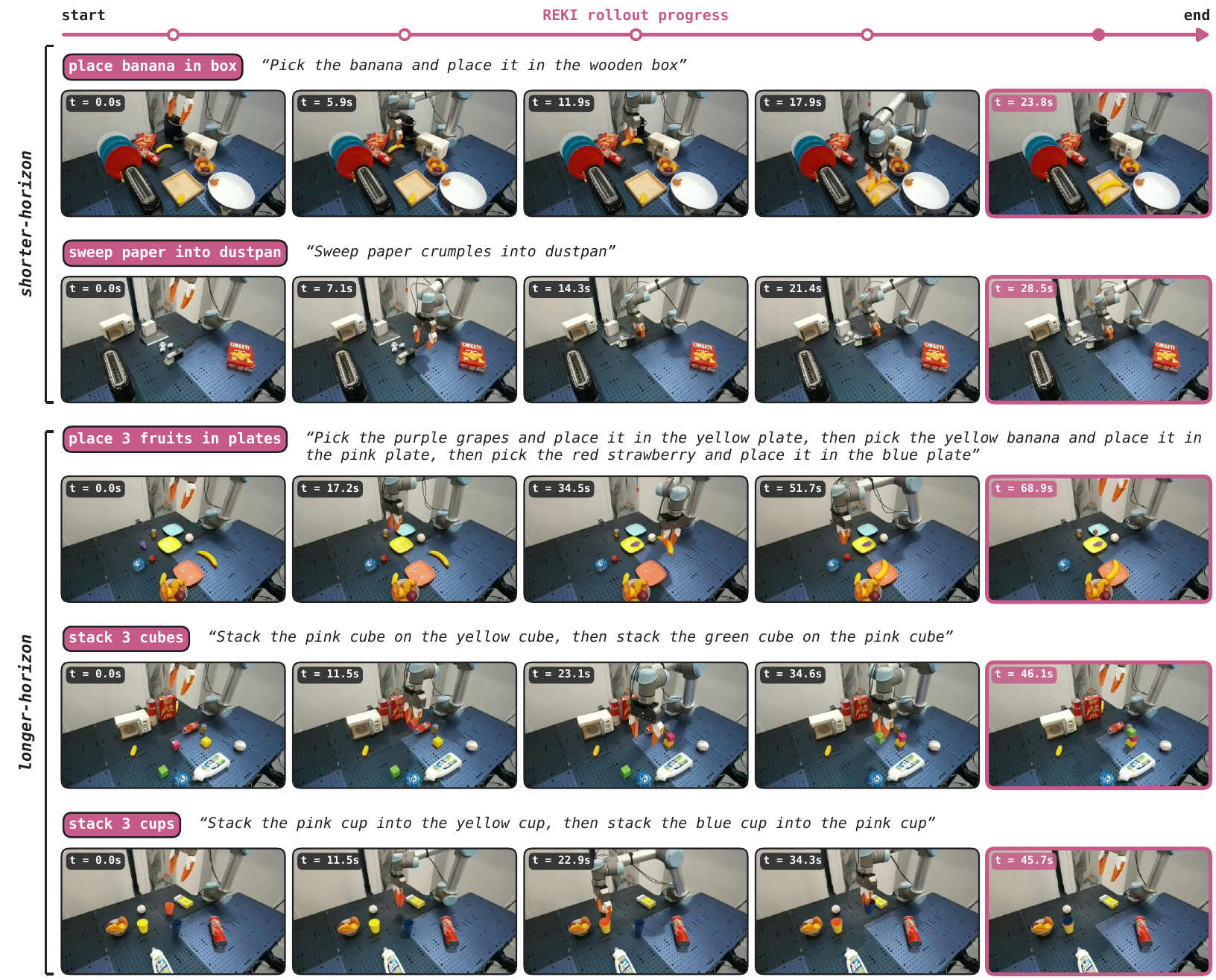}
\caption{\textbf{Qualitative examples.} Rollouts of REKI on our UR5 robotic arm across the five manipulation tasks.}
\label{fig:real-robot-qualitative}
\end{figure}

\Cref{fig:real-robot-qualitative} shows one rollout of REKI on each task, from the start to the end of the episode.

\end{document}